\pdfoutput=1

\documentclass[11pt]{article}

\usepackage{acl}

\usepackage{times}
\usepackage{latexsym}

\usepackage[T1]{fontenc}

\usepackage[utf8]{inputenc}

\usepackage{microtype}
\usepackage{times}
\usepackage{latexsym}
\usepackage{graphicx}
\usepackage{booktabs}
\usepackage{multirow}
\usepackage{float}
\usepackage{xcolor}
\usepackage{colortbl}
\usepackage{todonotes}
\usepackage{color}
\usepackage{tabularx}
\usepackage{array}
\usepackage{caption}
\usepackage{subcaption}
\usepackage{amsmath}
\usepackage{placeins}

\definecolor{darkgreen}{RGB}{0,100,0}

\title{Hate Speech According to the Law: An 
Analysis for Effective Detection}

\author{Katerina Korre$^1$\and John Pavlopoulos$^2$\and Paolo Gajo$^1$ \and Alberto Barrón-Cedeño$^1$\\
$^1$DIT, University of Bologna, Forlì, Italy \\
   $^2$Department of Informatics, Athens University of Economics and Business, Athens, Greece\\
 $^1$\{aikaterini.korre2, paolo.gajo2, a.barron\}@unibo.it\\
 $^2$annis@aueb.gr}

\begin{document}
\maketitle
\begin{abstract}
The issue of hate speech extends beyond the confines of the online realm. 
It is a problem with real-life repercussions, prompting most nations to formulate legal frameworks that classify hate speech as a punishable offence. These legal frameworks differ from one country to another, contributing to the big chaos that online platforms have to face when addressing reported instances of hate speech. With the definitions of hate speech falling short in introducing a robust framework, we turn our gaze onto hate speech laws. We consult the opinion of legal experts on a hate speech dataset and we experiment by employing various approaches such as pretrained models both on hate speech and legal data, as well as exploiting two large language models (Qwen2-7B-Instruct and Meta-Llama-3-70B). Due to the time-consuming nature of data acquisition for prosecutable hate speech, we use pseudo-labeling to improve our pretrained models. This study highlights the importance of amplifying research on prosecutable hate speech and provides insights into effective strategies for combating hate speech within the parameters of legal frameworks. Our findings show that legal knowledge in the form of annotations can be useful when classifying prosecutable hate speech, yet more focus should be paid on the differences between the laws.
\end{abstract}

\noindent
\textit{\textbf{Warning:} This paper contains hate speech that might be triggering for some individuals.}

\section{Introduction}
\label{sec:intro}

The detection of online hate speech is a growing challenge, not only due to the pressing need to address this issue but also due to the intricate and multifaceted nature of hate speech as a linguistic, social, and legal topic~\cite{JAHAN2023126232}. The possibility to interact online has enabled diverse groups to get involved in media governance, facilitating the spread of harassment, intimidation and stalking~\cite{Flick_2020}. The challenges researchers have to face are many, spanning both linguistic and computational realms, with the former encompassing issues like ambiguity~\cite{marwick2014online}, lack of context~\cite{markov-daelemans-2022-role,Perez}, multilinguality and multicultural variation~\cite{lee-etal-2023-hate}, while on the computational front the issues revolve around data bias~\cite{sap-etal-2019-risk}, imbalanced datasets~\cite{rathpisey19}, explainability~\cite{yang-etal-2023-hare}, and scalability~\cite{Yin2021}. Amidst these two categories lie significant legal and ethical concerns, especially since the legal landscape has struggled to keep pace with the rapid advancements in technology~\cite{aspray23}. 

Core to the aforementioned challenges is also the universal inability to agree on a single definition of hate speech~\cite{pachinger-etal-2023-toward,korre-etal-2023-harmful,khurana-etal-2022-hate, fortuna-etal-2020-toxic}, obstructing the creation of generalisable solutions both regarding legislations and NLP applications. The specific issue stems from the subjectivity that lies in the interpretation of hate speech due to different cultural or individual life experiences~\cite{waseem-2016-racist}. Especially in today's multicultural societies, it is necessary to be aware ``of the contexts, and of the power relations involved in balancing hate/free speech, as it is crucial in the analysis of regulation, legislative or other, as well as in any normative debate''~\cite{Maussen14}.

This paper addresses critical gaps in hate speech detection, that pertain to the lack of cultural awareness, as NLP must serve the speakers of a wide variety of languages, who come from a wide variety of cultures~\cite{hershcovich-etal-2022-challenges}, and the difficulty in establishing a universal definition of hate speech. Given that definitions can be subjective, we turn our attention to hate speech laws. Our research questions are: (1) How does the annotation of hate speech using laws, as compared to definitions, influence expert inter-annotator agreement, and does this effect vary across countries with different hate speech legislation? (2) Are the discrepancies in inter-annotator agreement reflected in the performance of pre-trained language models and large language models in hate speech detection? (3) Given the challenging and time-consuming nature of manually creating datasets for prosecutable hate speech, can LLM-generated instances be used to improve the performance of PLMs in detecting prosecutable hate speech?
Our annotation experiment uses three hate speech legislations from distinct countries, specifically Greece, Italy, and the United Kingdom.  
For our model experiments, we use
diverse pretrained BERT models (PLMs), while we also exploiting zero-shot and few-shot methods with two large language models (LLMs): Qwen2-7B-Instruct \cite{yang2024qwen2technicalreport} and Meta-Llama-3-70B-2-9b \cite{llama3modelcard}. Subsequently, we use Qwen2 to automatically generate labels for an additional 1000 instances, to be used to improve the performance of the PLMs. 
Our contributions can be summarized as follows:
\begin{itemize}
    \item We publish the first cross-country, expert-annotated dataset on prosecutable hate speech, considering the \textbf{legal ramifications} of published hate speech content and examining its legal compliance. 

    \item Through a \textbf{cross-country analysis} of hate speech, we demonstrate that inter-annotator agreement varies based on the legislation used during annotation. Interestingly, this variation does not necessarily overlap with the model evaluation, suggesting that PLMs and LLMs may carry their own inherent biases.

    \item \textbf{LLM-generated data }do not improve PLM performance, highlighting the still-present necessity of human annotation or a human-in-the-loop approach for prosecutable hate speech detection.
    
 \end{itemize}

\section{Related Work}
\label{sec:background}

\citet{GuillénNieto+2023} explains that hate speech as a social and legal notion are inherently connected. Socially, hate speech contributes to conflicts and conditions favorable to hate crimes. Legally, it is seen as ``an abstract endangerment statute'', balancing the rights to freedom of expression and dignity.

Hate speech on online platforms often precedes or coordinates racially motivated attacks.\footnote{https://www.cfr.org/backgrounder/hate-speech-social-media-global-comparisons}  For instance, Joshua Bonehill-Paine was prosecuted for racially harassing an ex-Labour MP.\footnote{https://www.theguardian.com/uk-news/2016/dec/07/racist-troll-guilty-harassing-labour-mp-luciana-berger-joshua-bonehill-paine} Given the detrimental effects of hate speech, a robust legal framework is needed. Incorporating NLP techniques with legal information can automate the detection of content with legal implications.

\subsection{Legal Frameworks on Hate Speech}
\label{sec:legalframeworks}
Forming legal frameworks for hate speech is a challenging task as legislation differs from one country to another. According to~\citet{marwick2014online} there are three main legal scholarly approaches to the definition of hate speech: (i) \textit{content-based}, which includes words, expressions, symbols and iconographies generally considered offensive to a particular group of people and objectively offensive to society; (ii) \textit{intent-based}, which requires the speaker’s communicative intention to incite hatred or violence against a particular minority, member of a minority, or person associated with a minority without communicating any legitimate message; (iii) and \textit{harms-based}, that causes the victim harm, such as loss of self-esteem, physical and mental stress, social and economic subordination and effective exclusion from mainstream society. These definitions share at least one core element with one another, which hints the potential for a unified framework. 

A lot of steps have been made in the European Union (EU) in the past, such as the \textit{No Hate Speech Movement} by the Council Of Europe,\footnote{https://www.coe.int/en/web/no-hate-campaign/no-hate-speech-movement} a campaign that mobilises young people to report hate speech and cyberbullying to the relevant authorities and on social media channels, and the \textit{EU Hate Speech Code}, which required social media platforms to review most notifications potentially flagged for hate speech within 24 hours.\footnote{https://www.theguardian.com/technology/2016/may/31/\\facebook-youtube-twitter-microsoft-eu-hate-speech-code}
After long negotiations, the EU has managed to set up a common regime that requires that all EU Member States make incitement to hatred or violence a punishable act. This is called \textit{The Council Framework Decision} and it will come into force in 2024.\footnote{Council Framework Decision 2008/913/JHA of 28 November 2008 on combating certain forms and expressions of racism and xenophobia by means of criminal law. In the remainder of this paper, we shall refer to this as ‘EU law’ or ‘EU Framework Decision’ for simpliﬁcation.} The EU Code of Conduct and the EU’s new Digital Services Act require also platforms, such as Twitter and Facebook, to take action with regard to illegal content,\footnote{Regulation (EU) 2022/2065 of the European Parliament and of the Council of 19 October 2022 on a Single Market For Digital Services and amending Directive 2000/31/EC (Digital Services Act).}  while in 2018, the introduction of the Audiovisual Media Directive called upon member states to guarantee that audiovisual media services, provided by media service providers and video sharing platform providers under their jurisdiction, to not include any instigation of violence or hostility directed at any group or individual based on the criteria listed in Article 21 of the EU Charter of Fundamental Rights, which encompasses gender as well~\cite{Flick_2020}.\footnote{Directive (EU) 2018/1808 of the European Parliament and of the Council concerning the provision of audiovisual media services (Audiovisual Media Services Directive).}

\subsection{Legal NLP on Hate Speech}
One of the few attempts to operationalize the decision as to whether a post is subject to criminal law was conducted by~\citet{zufall-etal-2022-legal}, who translated the legal framework as deﬁned in the EU Framework Decision 2008/913/JHA into a sequence of binary decisions. Their results suggest that annotation is more efficient when breaking the decision  into two subtasks: target group detection (which groups are referred to in a post, for example minorities, such as women and immigrants) and conduct detection (what punishable action patterns are referred to in a post), enhancing the transparency and explainability compared to black-box models. For their purposes, they create a Punishable Hate Speech Dataset in German, with a moderate inter-annotator agreement (IAA) for target group and targeting conduct (Cohen's Kappa 0.52 to 0.70), but low for the punishable label (0.33 to 0.43). Another attempt to exploring hate speech laws from an NLP outlook was that of~\citet{fiser-etal-2017-legal}, which presented the legal framework, a dataset, as well as an annotation schema for unacceptable discourse practices in Slovene, which includes annotations for the typology and target of the unacceptable discourse.%
\footnote{Unfortunately, neither of these two datasets are publicly available due to privacy issues, to comply with the General Data Protection Regulation (GDPR). 
\url{https://commission.europa.eu/law/law-topic/data-protection/data-protection-eu_en}}
\citet{khurana-etal-2022-hate} propose integrating legal insights into hate speech definitions, offering criteria drawn from both legal and social science perspectives. These criteria aim to enhance the precision of hate speech definitions and annotation guidelines across five dimensions: (1) target groups, (2) dominance, (3) perpetrator characteristics, (4) type of negative group reference, and (5) potential consequences/effects. By presenting this framework, researchers can better select suitable datasets, either directly applicable or as foundational resources with additional specifications. Compared to~\citet{khurana-etal-2022-hate} and~\citet{zufall-etal-2022-legal}, we use the entire legislation, as our main focus is the differences between the various hate speech laws, and we preserve a low inter-annotator agreement, to highlight and to examine those differences both in terms of human judgement and AI models. More recently, \citet{luo-etal-2023-legally} introduced hate speech detection grounded in enforceable legal definitions. They create a gold standard dataset annotated by legal experts and establish baseline
LLMs, incorporating explanations. Although these models demonstrate strong performance in quantitative metrics, a closer examination of their explainability reveals that the explanations lack coherent reasoning. Compared to~\citet{luo-etal-2023-legally}, our study relies on laws from 3 different countries, while they extracted 11 \textbf{definitions} of hate speech from the Canadian Criminal Code, as well as the Human Rights Code, and various Hateful Conduct Policies. The labeling is also different; \citet{luo-etal-2023-legally} use ``Violates'', ``Does not Violate'' ``Unclear'', while in this study we use ``Prosecutable'', ``Likely Prosecutable'', ``Unlikely Prosecutable'', and ``Not Prosecutable''. Finally, their task formulation deems an instance positive (i.e., ``violates'') if it breaches any of the provided definitions, whereas our study assesses each national hate speech law individually, providing separate expert annotations accordingly.

\section{Method}
\label{sec:method}

In this study, we focus on the laws of Greece, Italy, and the UK, for prosecutable hate speech detection purposes. The laws are retrieved from the Global Handbook of Hate Speech Laws, which provides a translated version of the laws into English (see Appendix~\ref{sec:appendixA}).\footnote{https://futurefreespeech.com/global-handbook-on-hate-speech-laws/} Our method involves, (1)~obtaining expert judgements on a sample of HatEval on prosecutable hate speech; (2)~exploiting state-of-the-art hate speech PLMs, and LLMs, using zero-shot, few-shot techniques and leave-one-out cross validation (LOOCV); and (3)~generating silver data for on an additional 1000 instances for improving the PLMs.


\subsection{Data and Annotation}

For our annotation, we use instances from HatEval~\cite{basile-etal-2019-semeval}, a dataset in English which contains hate speech instances both against immigrants and women, annotated via crowdsourcing. 
A total of 100 of hateful instances were extracted from the dataset.

Our re-annotation was conducted with the help of three experts. Expert A is a legal expert, currently a PhD candidate in Criminal law. Expert B is a master's student in Criminology. Expert C  holds a master degree in European Law. All of them possess a near-native level of English. They all annotated the sentences voluntarily after examining the three laws (Greek, Italian, and UK) by providing labels about whether the given instances were \textbf{prosecutable} hate speech or not. There was also a section for comments that allowed us to detect some ambiguous or controversial cases. Each annotation round, which involved evaluating the 100 examples, took around 10 hours to complete for all three laws. The experts reported that they needed 3 hours to read and study the document on the laws and make a summary. They reported that they had to study additional legislative material to be able to judge responsibly. Judging the instances took about 7 hours in total.
According to the two experts, this is not an easy task.
Some tweets were easy to understand and assessing whether the hate speech in them could be subject to prosecution was straightforward. Some others required a significant amount of 
research, which involved the interpretation of the meaning of the hashtags and abbreviations, as well as the disambiguation of slang and ambiguous words according to the interpreted intent and context of the post. For example, there were cases where the instance could be liable to be sued in court but only for insult or defamation rather than hate speech.
This type of annotation is not feasible by other means, such as crowdsourcing.

\begin{table}[!tb]
\centering
\small
\setlength{\tabcolsep}{2pt} 
\begin{tabular}{@{}l|ccc|cccc@{}}
\toprule
\textbf{National}  & \multicolumn{3}{c}{\textbf{Cohen's kappa}} & \multicolumn{4}{c}{\bf Class: is prosecutable?}  \\
\textbf{Law} & \textbf{A \& B} & \textbf{B \& C} & \textbf{C \& A} & \textbf{Yes} & \textbf{Likely} & \textbf{Unlikely} & \textbf{No}\\
\midrule
Greece & 0.20 & 0.25 & 0.12 & 4 & 16 & 32 & 48\\
Italy  & 0.31 & 0.59 & 0.37 & 9 & 16 & 17 & 58\\
UK     & 0.39 & 0.61 & 0.30 & 8 & 11 & 22 & 59\\
\bottomrule
\end{tabular}
\caption{Cohen's kappa and the number of instances per ordinal class according to the different country laws.}
\label{tab:iaa}
\end{table}

Table~\ref{tab:iaa} reports the agreement measures. The three experts labeled the instances differently, with Experts B and C being the pair with the highest agreement (0.59 for Italy and 0.61 for the UK). The poor to slight agreement among the pairs of experts suggests inconsistency in their assessments of whether something is prosecutable according to the Greek law, with an average agreement of 0.19. For the Italian and the UK laws, the agreement is moderate, with an average agreement of 0.42 and 0.43, respectively.
Hate speech datasets typically present low inter-annotator agreement, due to the subjective nature of the task as well as its dependance on the annotation scheme \cite{Casula2020}.
The results also point to the subjectivity of law interpretation ---even by experts--- a frequent phenomenon that, among others, often leads to the appealing of court decisions~\citep{dworkin1988subjectivity}. The challenge of achieving a high inter-annotator agreement is explained in~\citet{braun2023i} and in~\citet{wacholder-etal-2014-annotating} where they emphasize the inherent subjectivity of legal annotation, as each individual interprets the law in a different manner, adding that in many cases the inter-annotator agreement is not even reported.


The judgement variation relates to the main criterion for considering an instance of hate speech liable under the law, which is the fact that the offensive or aggressive verbal act must \textbf{incite violence} on the basis of a social or minority characteristic, e.g., gender, religion, race. In that case, and due to the lack of context and other verbal markers, it proved hard for the experts to clearly perceive the possibility of whether those instances could in fact incite violence. 

Additionally, the disagreement may arise from the subjective global knowledge or experiences of each expert. For example, in the case of the tweet, ``Deport them all! \#BuildThatWall'', two experts felt that the wording and call for deportation did not necessarily reflect racial motives. However, the third expert, taking into account factors such as the timing of the tweet, the political climate, and the implied reference to the US--Mexico border wall, interpreted the tweet as prosecutable hate speech. These differing perspectives do not necessarily imply that one interpretation is wrong. Instead, it highlights the need for clear guidelines in the task—specifically, whether background information, such as political context, should be considered or not. In summary, the disagreement appears to stem from varying perspectives on issues like the potential for incitement to violence, the distinction between defamation and hate speech, and how one connects the instance to current events. 

Taking this to consideration, we align with the opinion of~\citet{Basile2020ItsTE} who claims that when it comes to annotating highly subjective tasks such as hate speech, the opinions of all the experts can be correct and that disagreeing annotations that come from diverging opinions should be equally considered in the construction of a gold standard dataset. Due to those differences, and with respect to the opinions of all the experts, we decided to formulate the task as an ordinal regression problem. We consider cases where all experts label positive as \textbf{prosecutable hate speech}, cases where two experts agree that an instance is prosecutable as \textbf{likely prosecutable}, cases where only one expert claims that an instance is prosecutable as \textbf{unlikely prosecutable}. Finally, cases where all experts label negative as \textbf{not prosecutable hate speech}. Table~\ref{tab:iaa} presents the class distribution.\footnote{We plan to release our data upon acceptance.}


By examining texts labeled differently by the annotators,
we also observe differences and similarities in the legal principles. For instance, the Italian hate speech law (Law 167/2017) overlooks gender issues, resulting in most instances being outside the scope of prosecutability, whereas they might be punishable under the laws of Greece and the UK. On the other hand, there is a gap in interpretation of legal coverage against hateful language towards immigrants in Greek and Italian legislation, where many cases considered negative are deemed positive according to UK law.



\subsection{PLMs} 
\label{baselines}
Using our legal expert-judged dataset, we fine-tune and test four BERT pretrained models~\citep{devlin-etal-2019-bert}. For their assessment, we 
depart from three hate speech BERT models and one legal BERT model:


\paragraph{HateBERT.}~\citep{caselli-etal-2021-hatebert} based on the English BERT base model~\citep{devlin-etal-2019-bert}. It was further trained for the task of hate speech detection on more than 1\,M posts from banned communities from Reddit. 

\paragraph{DehateBERT.}~\citep{aluru2020deep} finetuned on multilingual BERT\footnote{\url{https://github.com/google-research/bert/blob/master/multilingual.md}} and trained for the task of hate speech detection with different learning rates in a monolingual English setting, using various established hate speech datasets.

\paragraph{HateRoBERTa.}~\citep{vidgen2021lftw}  which used a human-and-model-in-the-loop process for training an online hate detection system using RoBERTa~\cite{liu2019roberta} to dynamically  create hate speech datasets.

\paragraph{LegalBERT.}~\cite{chalkidis-etal-2020-legal} which is pretrained on 12 GB of diverse English legal text from several fields (e.g., legislation, court cases, contracts) scraped from publicly available resources, using BERT-base~\citep{devlin-etal-2019-bert}.

Our assessment of the capabilities of the models is carried out on the basis of
LOOCV. That is, we perform 100 train-and-testing runs to have all instances as the test set once.
In all cases, we fine-tune the models for 50 epochs, with early stopping with patience 2, with a learning rate of 1e-5, an AdamW optimizer, and a batch size of 32. Our output layer has four units corresponding to the four classes, and we use a softmax activation function. We select the class with the highest probability for the prediction.
The classes are mapped accordingly:
\[
\text{output} =
\begin{cases}
    \text{non prosecutable} & \text{if } \text{class} = 0\\
    \text{unlikely prosecutable} & \text{if } \text{class} = 1\\
    \text{likely prosecutable} & \text{if } \text{class} = 2\\
    \text{prosecutable} & \text{if } \text{class} = 3
\end{cases}
\]

We evaluate our models using Mean Absolute Error (MAE) and Mean Squared Error (MSE), with a focus on the differences between the models and the accuracy of their predictions compared to the gold-standard. A smaller error value indicates better model performance, with a maximum MAE of 3 and a maximum MSE of 9. Additional F$_1$ scores are provided in Appendix~\ref{appendixD}.

\subsection{LLMs}
We also experiment with two LLMs and six prompting strategies, to extend our benchmark while investigating possible biases of the LLMs in favour of laws of a specific country. We use Qwen2-7B-Instruct \cite{yang2024qwen2technicalreport} and Meta-Llama-3-70B-2-9b \cite{llama3modelcard}. 
We consider both zero- and few-shot settings:

\paragraph{Zero-shot agnostic (0-shot).} The prompt asks to what degree a given instance should be considered prosecutable hate speech. In this case, we do not include any example (be it positive or negative) from our dataset, nor a legislation, in the prompt.

\paragraph{Zero-shot with law (0-shot w/Law).} We do the same as in 0-shot, but this time we also provide the full country-specific hate speech law in the prompt. 

\paragraph{Few-shot agnostic ($k$-shot).}As in 0-shot, but this time we concatenate a balanced amount of $k$ randomly-selected instances to the prompt.

\paragraph{Few-shot with law ($k$-shot w/Law).} As in 0-shot w/Law, but this time we concatenate a balanced amount of $k$ randomly-selected instances to the prompt.

\paragraph{LOOCV agnostic.} As in the $k$-shot, but this time we concatenate 99 instances selected from the dataset, leaving out one instance each time for validation, and use the remaining instances as the prompt.

\paragraph{LOOCV with law.} As in the LOOCV, but this time we also concatenate the law to the prompt.

In the few-shot settings, we use $k\in \{4, 8, 12\}$.


\section{Experimental Results}
We present our experimental results in three segments, beginning with the benchmarking of the PLMs, followed by an analysis of the two LLMs, and concluding with a discussion on the impact of silver labels on the PLMs' performances.
\label{sec:results}

\subsection{PLM Leave-One-Out Cross-validation}

Table~\ref{tab:baselines} provides a comparative analysis of the different models' performance in predicting the ordinal values representing prosecutable hate speech degrees across the three different country laws.

DehateBERT consistently demonstrates the highest accuracy for the laws of Greece, achieving the lowest MAE (0.54) and MSE (0.68), indicating its robustness in this context. For Italy, LegalBERT performs best, with the lowest MAE (0.44) and MSE (0.62), showing that it is more adept to handle Italian texts. HateRoBERTa performs better with the UK-law-annotated part of the dataset, with the lowest MAE (0.50), while LegalBERT shows the overall best MSE (0.80) for the UK. This variation among the models highlights the importance of evaluating them within specific linguistic, cultural, and legal contexts to ensure their effectiveness. Overall, selecting models like DehateBERT for Greece, LegalBERT for Italy, or HateRoBERTa for the UK can lead to more accurate and reliable outcomes in prosecutable hate speech detection tasks for these countries.

\subsection{LLM Prompting}
Tables~\ref{tab:zero_llms} and \ref{tab:fewshots_llms} provide insights into the LLMs' performance with different prompting setups. We observe that in the majority of the cases the error scores drop when we integrate the actual legislation in the prompt. In some cases the performance is improved the more examples we use, as we see with Qwen2, in the 12-shot and the LOOCV setups. Although the error scores are low, the overall performance is lacking (evident also by looking at the F$_1$-scores in Appendix~\ref{appendixD}), as there are cases where one of the classes is frequently completely ignored by the LLM. This does not happen with the multiclass PLM models.  

When comparing the two models, Qwen2 outperforms Llama3, which might hint it already contains more legal knowledge incorporated during training.
With regard to which law-annotated dataset allows better prediction when incorporated in these models, in the 12-shot and the LOOCV, we see that the best scores are achieved when using the Greek and UK law-annotated datasets, except for the few-shot setting with Qwen2, where the best scores are achieved with the Italian law-annotated datasets. Yet, the Greek and UK laws are also more comprehensive on a natural language level since they contain more details about the possible targeted minorities and the potential punishable hate speech contexts, as noted by the experts in Section \ref{sec:method} (see also Appendix~\ref{sec:appendixA}).

\begin{table}[!tb]
    \centering
    \resizebox{\columnwidth}{!}{%
    \begin{tabular}{lrrrrrr}
    \toprule
     & \multicolumn{2}{c}{\textbf{Greece}} & \multicolumn{2}{c}{\textbf{Italy}} & \multicolumn{2}{c}{\textbf{UK}}  \\
     \textbf{Model}  & MAE & MSE & MAE & MSE & MAE & MSE \\ \midrule
     HateBERT & 0.60 & \cellcolor[gray]{0.7}0.92 & \cellcolor[gray]{0.7}0.58 & 1.12 & 0.59 & 1.19 \\
     DehateBERT &\textbf{0.54} & \textbf{0.68} & \cellcolor[gray]{0.7}0.48 & \cellcolor[gray]{0.7}0.64 & 0.59 & 0.89 \\
     HateRoBERTa & 0.56 & \cellcolor[gray]{0.7}0.88 & 0.51 & 1.01 & \cellcolor[gray]{0.7}0.50 & 0.98 \\
     LegalBERT & 0.61 & 0.85 & \cellcolor[gray]{0.7}\textbf{0.44} & \cellcolor[gray]{0.7}\textbf{0.62} & \textbf{0.48} & \textbf{0.80 }\\ 
     \hline
     Average & 0.57 & 0.83 & 0.50 & 0.84 & 0.54 & 0.96 \\
     \bottomrule
    \end{tabular}%
    }
    \caption{MAE and MSE of the PLMs fine-tuned and evaluated on LOOCV. In gray background is the best scores per model. In bold is the best score per country. }
    \label{tab:baselines}
\end{table}

\begin{table*}[t]
    \centering
    \small
    \begin{subtable}{\textwidth}
    \centering
    \setlength{\tabcolsep}{4pt}{
    \begin{tabular}{@{}lcccc|cccccc@{}}
       & \multicolumn{2}{c}{\textbf{0-shot}} & \multicolumn{2}{c}{\textbf{0-shot w/Law}} & \multicolumn{2}{c}{\textbf{LOOCV few-shot}} & \multicolumn{2}{c}{\textbf{LOOCV few-shot w/Law}} \\ \cmidrule(lr){2-3} \cmidrule(lr){4-5} \cmidrule(lr){6-7} \cmidrule(lr){8-9}  
    \textbf{Country}   & MAE & MSE & MAE & MSE & MAE & MSE & MAE & MSE \\
     \hline
       Greece  &  \cellcolor[gray]{0.7}1.26 &  \cellcolor[gray]{0.7}2.54 & 1.24 & \cellcolor[gray]{0.7}2.08 &  \cellcolor[gray]{0.7}0.67 &  \cellcolor[gray]{0.7}1.11 & \cellcolor[gray]{0.7}\textbf{0.58} &  \cellcolor[gray]{0.7}\textbf{0.86}\\
       Italy & 1.33 & 2.87 & \cellcolor[gray]{0.7}1.07 & 2.27 & \textbf{0.71} & 1.45 & 0.73 & \textbf{1.41}\\
       UK & 1.31 & 2.71 &  1.51 &  3.49 & \cellcolor[gray]{0.7}\textbf{0.67} & \textbf{1.35 }& 0.81 & 1.71\\
    \end{tabular}%
    }

    \caption{Qwen2}    
    \label{tab:zero_qwen} 
    \end{subtable}

     
    \begin{subtable}{\textwidth}
    \centering
    \setlength{\tabcolsep}{4pt}{ 
    \begin{tabular}{@{}lcccc|cccccc@{}}
       & \multicolumn{2}{c}{\textbf{0-shot}} & \multicolumn{2}{c}{\textbf{0-shot w/Law}} & \multicolumn{2}{c}{\textbf{LOOCV few-shot}} & \multicolumn{2}{c}{\textbf{LOOCV few-shot w/Law}} \\ \cmidrule(lr){2-3} \cmidrule(lr){4-5} \cmidrule(lr){6-7} \cmidrule(lr){8-9} 
    \textbf{Country}   & MAE & MSE & MAE & MSE & MAE & MSE & MAE & MSE \\
     \hline
       Greece  &  \cellcolor[gray]{0.7}1.58 & \cellcolor[gray]{0.7}3.74  & 1.74 & 4.10 & \cellcolor[gray]{0.7}\textbf{0.94} &  \cellcolor[gray]{0.7}\textbf{1.44 }&  \cellcolor[gray]{0.7}1.13 &  \cellcolor[gray]{0.7}1.89\\
       Italy & 1.84 & 4.52 & 1.94 & 5.02 & \textbf{1.00} & \textbf{1.84} & 1.18 & 2.36\\
       UK & 1.72 & 4.30 & \cellcolor[gray]{0.7}1.42 & \cellcolor[gray]{0.7}3.30  & \textbf{1.30} & 3.04 &  1.33 & \textbf{3.03}\\
    \end{tabular}%
    }
    \caption{Llama3}    
    \label{tab:zero_llama3} 
    \end{subtable}

    \caption{MAE and MSE for the 0-shot and LOOCV approaches. The columns ``w/Law'' indicate that we also use the actual hate speech law in the prompt. In gray background is the best scores per setting. In bold is the best score per country.}
\label{tab:zero_llms} 
\end{table*}

\begin{table*}[!tb]
\small
    \centering
    \begin{subtable}{\textwidth}
    \centering
    \begin{tabular}{@{}lccccccccccccccc@{}}
                    & \multicolumn{6}{c}{\textbf{w/o Law}} & \multicolumn{6}{c}{\textbf{w/Law}} \\
                    \cmidrule(lr){2-7} \cmidrule(lr){8-13}
                    & \multicolumn{2}{c}{\textbf{4-shot}} & \multicolumn{2}{c}{\textbf{8-shot}} & \multicolumn{2}{c}{\textbf{12-shot}} 
                    & \multicolumn{2}{c}{\textbf{4-shot}} & \multicolumn{2}{c}{\textbf{8-shot}} & \multicolumn{2}{c}{\textbf{12-shot}} \\
                    \cmidrule(lr){2-3} \cmidrule(lr){4-5} \cmidrule(lr){6-7} 
                    \cmidrule(lr){8-9} \cmidrule(lr){10-11} \cmidrule(lr){12-13}
                    \textbf{Country} & MAE & MSE & MAE & MSE & MAE & MSE & MAE & MSE & MAE & MSE & MAE & MSE\\
                    \midrule
                    Greece &  0.91 & \cellcolor[gray]{0.7}1.53 & 0.89 & \cellcolor[gray]{0.7}1.61 & \textbf{0.87} & \textbf{1.47} & \textbf{0.78} & \cellcolor[gray]{0.7}\textbf{1.22} & 0.84 & 1.42 & 0.91 & 1.57 \\
                    Italy  &  0.92 & 1.68 & \cellcolor[gray]{0.7}0.88 & 1.72 & \cellcolor[gray]{0.7}\textbf{0.67} & \cellcolor[gray]{0.7}\textbf{1.27} & \cellcolor[gray]{0.7}0.58 & 1.98 & \cellcolor[gray]{0.7}0.63 & \cellcolor[gray]{0.7}0.99 & \cellcolor[gray]{0.7}\textbf{0.57} & \cellcolor[gray]{0.7}0\textbf{.91} \\
                    UK     &  \cellcolor[gray]{0.7}\textbf{0.90} & \textbf{1.88} & 0.94 & 1.96 & 0.93 & 2.11 & 0.98 & 2.02 & 0.91 & 1.67 & \textbf{0.73} & \textbf{1.29}\\
                    \bottomrule
    \end{tabular}
    \caption{Qwen2}
    \label{tab:comparison_few_shots}
  \end{subtable}%
    \vspace{0.5cm} 
    \begin{subtable}{\textwidth}    
    \centering
    \begin{tabular}{@{}lccccccccccccccc@{}}
                    & \multicolumn{6}{c}{\textbf{w/o Law}} & \multicolumn{6}{c}{\textbf{w/ Law}} \\
                    \cmidrule(lr){2-7} \cmidrule(lr){8-13}
                    & \multicolumn{2}{c}{\textbf{4-shot}} & \multicolumn{2}{c}{\textbf{8-shot}} & \multicolumn{2}{c}{\textbf{12-shot}} 
                    & \multicolumn{2}{c}{\textbf{4-shot w/Law}} & \multicolumn{2}{c}{\textbf{8-shot w/Law}} & \multicolumn{2}{c}{\textbf{12-shot w/Law}} \\
                    \cmidrule(lr){2-3} \cmidrule(lr){4-5} \cmidrule(lr){6-7} 
                    \cmidrule(lr){8-9} \cmidrule(lr){10-11} \cmidrule(lr){12-13}
                    \textbf{Country} & MAE & MSE & MAE & MSE & MAE & MSE & MAE & MSE & MAE & MSE & MAE & MSE\\
                    \midrule
                    Greece &  \cellcolor[gray]{0.7}\textbf{0.90} & \cellcolor[gray]{0.7}\textbf{1.50 }& 1.21 & \cellcolor[gray]{0.7}2.47 & \cellcolor[gray]{0.7}1.02 & \cellcolor[gray]{0.7}1.94 & \cellcolor[gray]{0.7}\textbf{1.06} & \cellcolor[gray]{0.7}\textbf{1.88} & \cellcolor[gray]{0.7}1.11 & \cellcolor[gray]{0.7}2.23 & 1.16 & \cellcolor[gray]{0.7}2.14 \\
                    Italy  &  \textbf{1.08} & \textbf{2.06} & 1.25 & 2.63 & 1.35 & 3.03 & \textbf{1.08} & \textbf{2.04} & 1.21 & 2.47 & \cellcolor[gray]{0.7}1.12 & 2.20 \\
                    UK     &  \textbf{1.09} & \textbf{2.33} & \cellcolor[gray]{0.7}1.15 & 2.59 & 1.22 & 2.66 & \textbf{1.08 }& \textbf{2.18} & 1.18 & 2.46 & 1.15 & 2.19\\
                    \bottomrule
        \end{tabular}
        \caption{Llama3}
        \label{tab:comparison_few_shots_llama3}
    \end{subtable}
    \caption{MAE and MSE for the the few-shot approaches. The column ``w/Law'' indicates that we also use the actual hate speech law in the prompt. In gray background is the best scores per setting. In bold is the best score per country.}
    \label{tab:fewshots_llms} 
\end{table*}

\subsection{PLMs: Revisited}
\label{sec:use_case}

\begin{figure*}
    \centering
    \includegraphics[width=\textwidth]{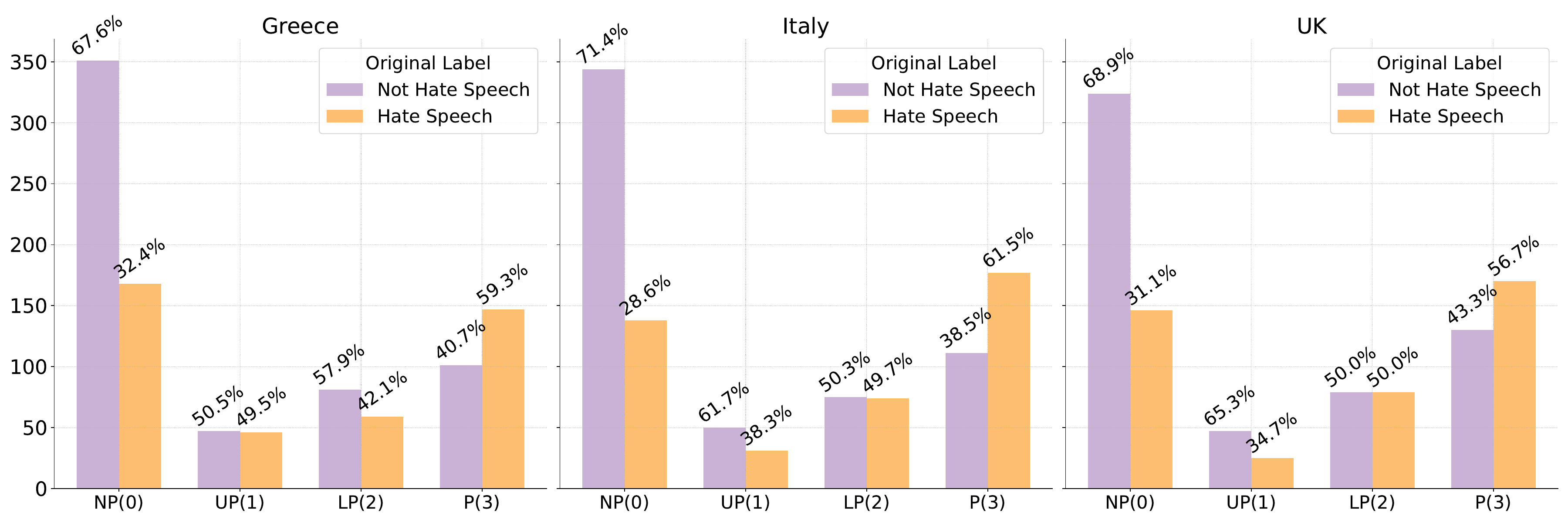}
    \caption{Distribution of classes as predicted on 1,000 instances from HateEval~\cite{basile-etal-2019-semeval}. The y axis presents the absolute number of instances. The labels correspond to NP for Not Prosecutable, UP for Unlikely Prosecutable, LP for Likely Prosecutable, and P for Prosecutable. The bar plots illustrate the original binary labels of the instances, distinguishing between hate speech and non-hate speech.}
    \label{fig:distribution_silver}
\end{figure*}

\begin{table}[t]
\centering
\small
\begin{tabularx}{\columnwidth}{@{}l|*{4}{>{\centering\arraybackslash}X}c@{}}
\toprule
\textbf{National Law} & \multicolumn{3}{c}{\textbf{Class: is prosecutable?}}\\
\textbf{} & \textbf{Yes} & \textbf{Likely} & \textbf{Unlikely} & \textbf{No}\\
\midrule
Greece & 147 & 59 & 47 & 351 \\
Italy & 177 & 74 & 50 & 344 \\
UK & 170 & 79 & 47 & 324 \\
\bottomrule
\end{tabularx}
\caption{Prosecutability of Silver Labels under National Laws: Number of cases classified as `No',` Potentially', and `Yes' for prosecution in Greece, Italy, and the UK.}
\label{tab:silver_stats}
\end{table}

Following our benchmark with PLMs and LLMs, we ask the following research question: ``Can LLM-generated instances be used to improve the performance of PLMs in prosecutable hate speech detection?''. To answer this question, we use our best performing LLM, i.e. Qwen2, to predict on 1,000 instances of HateEval dataset, creating unseen silver labels. In this way, we are also able to examine any bias towards any of the country laws the LLM might introduce to the PLMs.

Figure~\ref{fig:distribution_silver} illustrates the distribution of our model's prosecutable predictions when aligned with the original hate speech labels. Contrary to what was observed during manual annotation, the `Not prosecutable' label is still the most prevalent, however, it is followed by the `Prosecutable label', then 'Likely prosecutable', and with the `Unlikely prosecutable' label being the least frequent. From this, it can be inferred that there could be a significant number of prosecutable cases for instances that are originally judged simply as hate speech in available hate speech datasets. Examining the instances identified as `Not prosecutable' by the model, reveals notable model inaccuracies, more specifically, a considerable rate of false positives (31\% for Greece, 32\% for Italy, and 36\% for the UK). Analyzing this contrast between hate speech and prosecutable hate speech enables us to filter our silver dataset by eliminating instances of false positives and false negatives. Detailed class statistics for the finalized silver dataset are provided in Table~\ref{tab:silver_stats}.

We fine-tune the PLMs with a similar setup as in Section~\ref{sec:method} but this time we use the initial 100 instances exclusively as our test set. Our results are illustrated in Table~\ref{tab:baselines_revisited}. At a first glance, there is a significant deterioration with regard to predicting the prosecutability of the instances with the pretrained models. Upon manual examination and looking at the per class F$_1$ scores in the Appendix (Table~\ref{tab:plms_f1_2}), we observe that most models consistently miss to assign the label `Likely prosecutable' or  `Unlikely prosecutable', with the majority labeling the instance as `not prosecutable' or 'prosecutable'. This could be due to a variety of factors, including the complexity and ambiguity of the task, the distribution of labels within the dataset, or inherent uncertainties in the input data. 
Going back to our research question, LLM-generated data do not improve the overall scores. 


\begin{table}[!tb]
    \centering
    \resizebox{\columnwidth}{!}{%
    \begin{tabular}{lrrrrrr}
    \toprule
     & \multicolumn{2}{c}{\textbf{Greece}} & \multicolumn{2}{c}{\textbf{Italy}} & \multicolumn{2}{c}{\textbf{UK}}  \\
     \textbf{Model}  & MAE & MSE & MAE & MSE & MAE & MSE \\ \midrule
     HateBERT & 1.30 & 2.94 & 1.90 & 5.04 & 1.71 & 4.47 \\
     DehateBERT & 1.92 & 4.48 & 1.96 & 5.22 & 1.93 & 5.13\\
     HateRoBERTa &  1.95 &   4.89 & 2.10 & 5.64 & 2.13 & 5.73\\
     LegalBERT &  1.00 & 2.06 & 1.13 & 2.73 & 0.79 & 1.71 \\ \bottomrule
    \end{tabular}%
    }
    \caption{MAE and MSE of the pretrained models fine-tuned on the filtered silver labels and evaluated on our expert-annotated dataset. 
    }
    \label{tab:baselines_revisited}
\end{table}

\subsection{Error Analysis}
To summarize our analysis, we perform an error analysis looking at which classes the errors occurred for each model and setting. The PLMs consistently exhibited low error scores, ranging from 0.50 to 0.60 MAE, with the majority of their errors concentrated in the middle categories (Unlikely prosecutable, Likely prosecutable). In comparison, Qwen2, under its best configuration (12-shot with law), achieved a MAE of 0.57. However, its errors were primarily observed in categories 0 and 1, indicating a shift in error distribution relative to the PLMs. Llama 3, on the other hand, showed even higher error rates, with a similar concentration of errors in categories 0 and 1. The confusion matrices of the predictions of each model can be found in Appendix~\ref{appendixE}.

Furthermore, fine-tuning PLMs with silver labels derived from the LLMs was found to increase error rates, suggesting that training with silver data may introduce more noise rather than improving model performance.

Given these results, it appears that pretrained multiclass models specifically designed for hate speech detection are more effective in the context of ordinal regression, as they likely incorporate relevant legal knowledge concerning hate speech.

\section{Conclusion}
\label{sec:conclusion}
This paper explored detecting prosecutable hate speech and the effectiveness of introducing legal knowledge during fine-tuning and prompting, using four PLMs and two LLMs. We present a newly judged sampled dataset on prosecutable hate speech and conduct experiments using the labeled data to prompt them to predict whether a given hate speech instance is prosecutable. Additionally, we use our Qwen2 to create silver data and evaluate on our re-annotated dataset on prosecutable hate speech. The results showed that the models struggle to grasp the nuances of these laws and they often fail to incorporate the specific legal information provided by the laws into their decision-making, relying more on the knowledge they gained during pretraining.  The models tend to be more conservative in predicting an instance as prosecutable compared to non-prosecutable. This caution might derive from understanding the legal constraints and requirements with regard to what constitutes prosecutable hate speech and what does not. This tendency mirrors the approach of human annotation, where both experts assigned labels similarly to the models, being generous with the ``Not Prosecutable'' class. Finally, our error analysis showed that multiclass PLMs trained using LOOCV can be more effective than LLMs for the task of prosecutable hate speech detection. 

In future work, we plan to expand the experiments with more LLMs, as well as use the laws of other countries to see which aspects of hate speech (i.e., gender, race, etc) can be perceived by the models. 

\section*{Limitations}
\label{Limitations}
This work has certain limitations. Due to the challenging nature of the annotation task, the re-annotated dataset comprises only 100 instances. Consequently, leveraging encoder models for training and validation poses significant challenges, as these models typically require larger datasets for optimal performance. Additionally, our analysis is restricted to the laws of three countries (Greece, Italy, and the UK). Including more countries would provide deeper insights into cultural differences within the laws. Finally, we used variations of a single prompt for all cases. Experimenting with diverse prompts could further optimize the results from the LLMs, potentially enhancing explainability.

\section*{Ethics Statement}
\paragraph{Annotation}
Before our dataset annotation process, we provided the volunteering experts with some knowledge about the task and its potential harm, and we have obtained their explicit written consent. The annotation has been assessed by the Ethics committee of our institution. Moreover, we acknowledge the inherent subjectivity involved in labeling data and recognize that the provision of specific labels does not inherently validate our methodology. Consequently, we understand the importance of further annotations to delve into the subjective nature of supervised tasks like ours.

\paragraph{The Task}
Similar to hate speech detection, prosecutable hate speech detection involves ethical concerns, such as the freedom of speech and privacy issues, as the balance between mitigating harm caused by hate speech and preserving individuals' rights to free expression, as well as their data privacy, can be a delicate task. involves significant ethical concerns, including freedom of speech and privacy issues. Balancing the mitigation of harm caused by hate speech with the preservation of individuals' rights to free expression and data privacy is a delicate task. The complexity is aggravated in the context of prosecutable hate speech due to its legal implications, which can greatly influence outcomes. Misinterpretation or misapplication of these standards carries the risk of wrongful accusations and legal actions, potentially causing harm to innocent individuals.

\section*{Working setting}
The experiments with the pretrained models are run on a Tesla T4 GPU, as well as a Quadro P4000 GPU. The time consumption of the experiments was around 30 to 40 minutes per fine-tuning.  

\section*{AI-assisted technologies}
The authors of this work used ChatGPT 3.5 to improve the language and readability of the paper. We take full responsibility of the content which has been properly reviewed and edited to reflect our own methods.

\bibliography{anthology,custom}

\appendix

\section{Laws}
\label{sec:appendixA}

In this appendix we reproduce the articles that describe prosecutable hate speech in the three countries . 

\subsection{Greece}
\label{sub:greece}
Law No 927/1797 on Punishing Acts or Activities aimed at Racial Discrimination. 

\paragraph{Article 1:} 
Incitement to violence or hatred.

\noindent
(1) “Anyone, who publicly incites, provokes, or stirs up, either orally or through the press, the Internet, or any other means, acts of violence or hatred against a person or group of persons or a member of such a group defined by reference to race, color, religion, descent or national or ethnic origin, sexual orientation, gender identity, or disability, in a manner that endangers the public order and exposes the life, physical integrity, and freedom of persons defined above to danger, will be punished by imprisonment of from three months to three years and a fine of Euros 5,000 to 20,000.

\noindent
(2) Anyone, who publicly incites, provokes, or stirs, either orally or through the press, the Internet, or any other means, acts of destruction against the assets of a person or group of persons defined by reference to race, color, religion, descent or national or ethnic origin, sexual orientation, gender identity, or disability, in a manner that endangers the public order and exposes the life, physical integrity, and freedom of persons defined above to danger, will be punished by imprisonment of from three months to three years and a fine of Euros 5,000 to 20,000.

\subsection{Italy}
\label{sub:italy}

Criminal code. Article 604 bis: Propaganda and incitement to commit crimes for reasons of racial, ethnic and religious discrimination. 

I. Unless the fact constitutes a more serious offence, the following are punished: a) with imprisonment of up to one year and six months or with a fine of up to 6,000 euros for propaganda based on ideas of superiority or ideas of racial or ethnic hatred, or propaganda which instigates to commit or commits acts of discrimination on the grounds of racial, ethnic, national or religious grounds; b) Imprisonment of six months to four years for a person who in any way, instigates to commit or commits violence or acts of provocation to violence for racial, ethnic, national or religious reasons.

Law 654/1975 on the Ratification and Execution of the International Convention on the Elimination of All Forms of Racial Discrimination: 

\paragraph{Article 3:} I. Unless the fact constitutes a more serious offence, the following are punished: a) with imprisonment of up to one year and six months or with a fine of up to 6,000 euros for propaganda based on ideas of superiority or ideas of racial or ethnic hatred, or propaganda which instigates to commit or commits acts of discrimination on the grounds of racial, ethnic, national or religious grounds; b) Imprisonment of six months to four years for a person who in any way, instigates to commit or commits violence or acts of provocation to violence for racial, ethnic, national or religious. Law 167/2017: Full Implementation of EU Framework Decision 2008/913/HAD This law, amongst others modifies Law 654/1975: The main relevant Italian Law is Law 205/1993 which makes it a crime to ``propagate ideas based on racial superiority or racial or ethnic hatred, or to instigate to commit or commit acts of discrimination for racial, ethnic, national or religious motives.'' The law also punishes those who ``instigate in any way or commit violence or acts of provocation to violence for racist, ethnic, national or religious motives.''

\subsection{UK}
\label{sub:uk}

Public Order Act 1986
Part III Racial Hatred

Acts intended or likely to stir up racial hatred
Section 18 Use of words or behaviour or display of written material.
(1) A person who uses threatening, abusive or insulting words or behaviour, or displays any written material which is threatening, abusive or insulting, is guilty of an offence if—
(a)he intends thereby to stir up racial hatred, or
(b)having regard to all the circumstances racial hatred is likely to be stirred up thereby.
(2) An offence under this section may be committed in a public or a private place, except that no offence is committed where the words or behaviour are used, or the written material is displayed, by a person inside a dwelling and are not heard or seen except by other persons in that or another dwelling.
(4) In proceedings for an offence under this section it is a defence for the accused to prove that he was inside a dwelling and had no reason to believe that the words or behaviour used, or the written material displayed, would be heard or seen by a person outside that or any other dwelling.
(5) A person who is not shown to have intended to stir up racial hatred is not guilty of an offence under this section if he did not intend his words or behaviour, or the written material, to be, and was not aware that it might be, threatening, abusive or insulting.
(6) This section does not apply to words or behaviour used, or written material displayed, solely for the purpose of being included in a programme [included in a programme service].
Section 19 Publishing or distributing written material.
(1)A person who publishes or distributes written material which is threatening, abusive or insulting is guilty of an offence if—
(a)he intends thereby to stir up racial hatred, or
(b)having regard to all the circumstances racial hatred is likely to be stirred up thereby.
(2) In proceedings for an offence under this section it is a defence for an accused who is not shown to have intended to stir up racial hatred to prove that he was not aware of the content of the material and did not suspect, and had no reason to suspect, that it was threatening, abusive or insulting.
(3) References in this Part to the publication or distribution of written material are to its publication or distribution to the public or a section of the public.
Section 20 Public performance of play.
(1) If a public performance of a play is given which involves the use of threatening, abusive or insulting words or behaviour, any person who presents or directs the performance is guilty of an offence if—
(a)he intends thereby to stir up racial hatred, or
(b)having regard to all the circumstances (and, in particular, taking the performance as a whole) racial hatred is likely to be stirred up thereby.
(2) If a person presenting or directing the performance is not shown to have intended to stir up racial hatred, it is a defence for him to prove—
(a) that he did not know and had no reason to suspect that the performance would involve the use of the offending words or behaviour, or
(b) that he did not know and had no reason to suspect that the offending words or behaviour were threatening, abusive or insulting, or
(c) that he did not know and had no reason to suspect that the circumstances in which the performance would be given would be such that racial hatred would be likely to be stirred up.
(3) This section does not apply to a performance given solely or primarily for one or more of the following purposes—
(a) rehearsal,
(b) making a recording of the performance, or
(c) enabling the performance to be [included in a programme service];
but if it is proved that the performance was attended by persons other than those directly connected with the giving of the performance or the doing in relation to it of the things mentioned in paragraph (b) or (c), the performance shall, unless the contrary is shown, be taken not to have been given solely or primarily for the purposes mentioned above.
(4) For the purposes of this section—
(a)a person shall not be treated as presenting a performance of a play by reason only of his taking part in it as a performer,
(b) a person taking part as a performer in a performance directed by another shall be treated as a person who directed the performance if without reasonable excuse he performs otherwise than in accordance with that person’s direction, and
(c) a person shall be taken to have directed a performance of a play given under his direction notwithstanding that he was not present during the performance;
and a person shall not be treated as aiding or abetting the commission of an offence under this section by reason only of his taking part in a performance as a performer.
(5) In this section “play” and “public performance” have the same meaning as in the Theatres Act 1968.
(6) The following provisions of the Theatres Act 1968 apply in relation to an offence under this section as they apply to an offence under section 2 of that Act—
section 9 (script as evidence of what was performed),
section 10 (power to make copies of script),
section 15 (powers of entry and inspection).
Section 21 Distributing, showing or playing a recording.
(1) A person who distributes, or shows or plays, a recording of visual images or sounds which are threatening, abusive or insulting is guilty of an offence if—
(a) he intends thereby to stir up racial hatred, or
(b) having regard to all the circumstances racial hatred is likely to be stirred up thereby.
(2) In this Part “recording” means any record from which visual images or sounds may, by any means, be reproduced; and references to the distribution, showing or playing of a recording are to its distribution, showing or playing of a recording are to its distribution, showing or playing to the public or a section of the public.
(3) In proceedings for an offence under this section it is a defence for an accused who is not shown to have intended to stir up racial hatred to prove that he was not aware of the content of the recording and did not suspect, and had no reason to suspect, that it was threatening, abusive or insulting.
(4) This section does not apply to the showing or playing of a recording solely for the purpose of enabling the recording to be included in a programme service.
Section 22 Broadcasting or including programme in cable programme service.
(1) If a programme involving threatening, abusive or insulting visual images or sounds is included in a programme service], each of the persons mentioned in subsection (2) is guilty of an offence if—
(a)he intends thereby to stir up racial hatred, or
(b)having regard to all the circumstances racial hatred is likely to be stirred up thereby.
(2) The persons are—
(a)the person providing the programme service,
(b)any person by whom the programme is produced or directed, and
(c)any person by whom offending words or behaviour are used.
(3) If the person providing the service, or a person by whom the programme was produced or directed, is not shown to have intended to stir up racial hatred, it is a defence for him to prove that—
(a)he did not know and had no reason to suspect that the programme would involve the offending material, and
(b)having regard to the circumstances in which the programme was [included in a programme service], it was not reasonably practicable for him to secure the removal of the material.
(4) It is a defence for a person by whom the programme was produced or directed who is not shown to have intended to stir up racial hatred to prove that he did not know and had no reason to suspect—
(a)that the programme would be [included in a programme service], or
(b)that the circumstances in which the programme would be so included would be such that racial hatred would be likely to be stirred up.
(5) It is a defence for a person by whom offending words or behaviour were used and who is not shown to have intended to stir up racial hatred to prove that he did not know and had no reason to suspect—
(a)that a programme involving the use of the offending material would be [included in a programme service], or
(b)that the circumstances in which a programme involving the use of the offending material would be so included, or in which a programme . . . so included would involve the use of the offending material, would be such that racial hatred would be likely to be stirred up.
(6) A person who is not shown to have intended to stir up racial hatred is not guilty of an offence under this section if he did not know, and had no reason to suspect, that the offending material was threatening, abusive or insulting.
Racially inflammatory material
Section 23 Possession of racially inflammatory material.
(1) A person who has in his possession written material which is threatening, abusive or insulting, or a recording of visual images or sounds which are threatening, abusive or insulting, with a view to—
(a)in the case of written material, its being displayed, published, distributed, [or included in a cable programme service], whether by himself or another, or
(b)in the case of a recording, its being distributed, shown, played, [or included in a cable programme service], whether by himself or another,
is guilty of an offence if he intends racial hatred to be stirred up thereby or, having regard to all the circumstances, racial hatred is likely to be stirred up thereby.
(2) For this purpose regard shall be had to such display, publication, distribution, showing, playing, [or inclusion in a programme service] as he has, or it may reasonably be inferred that he has, in view.
(3) In proceedings for an offence under this section it is a defence for an accused who is not shown to have intended to stir up racial hatred to prove that he was not aware of the content of the written material or recording and did not suspect, and had no reason to suspect, that it was threatening, abusive or insulting.
Part 3A Hatred against persons on religious grounds
Acts intended to stir up religious hatred
Section 29B: Use of words or behaviour or display of written material
(1) A person who uses threatening words or behaviour, or displays any written material which is threatening, is guilty of an offence if he intends thereby to stir up religious hatred.
(2) An offence under this section may be committed in a public or a private place, except that no offence is committed where the words or behaviour are used, or the written material is displayed, by a person inside a dwelling and are not heard or seen except by other persons in that or another dwelling.
(3) A constable may arrest without warrant anyone he reasonably suspects is committing an offence under this section.
(4) In proceedings for an offence under this section it is a defence for the accused to prove that he was inside a dwelling and had no reason to believe that the words or behaviour used, or the written material displayed, would be heard or seen by a person outside that or any other dwelling.
(5) This section does not apply to words or behaviour used, or written material displayed, solely for the purpose of being included in a programme service.
29C Publishing or distributing written material
(1) A person who publishes or distributes written material which is threatening is guilty of an offence if he intends thereby to stir up religious hatred.
(2) References in this Part to the publication or distribution of written material are to its publication or distribution to the public or a section of the public.
Section 29D: Public performance of play
(1) If a public performance of a play is given which involves the use of threatening words or behaviour, any person who presents or directs the performance is guilty of an offence if he intends thereby to stir up religious hatred.
(2) This section does not apply to a performance given solely or primarily for one or more of the following purposes—
(a)rehearsal,
(b)making a recording of the performance, or
(c)enabling the performance to be included in a programme service;
but if it is proved that the performance was attended by persons other than those directly connected with the giving of the performance or the doing in relation to it of the things mentioned in paragraph (b) or (c), the performance shall, unless the contrary is shown, be taken not to have been given solely or primarily for the purpose mentioned above.
(3) For the purposes of this section—
(a)a person shall not be treated as presenting a performance of a play by reason only of his taking part in it as a performer,
(b)a person taking part as a performer in a performance directed by another shall be treated as a person who directed the performance if without reasonable excuse he performs otherwise than in accordance with that person’s direction, and
(c)a person shall be taken to have directed a performance of a play given under his direction notwithstanding that he was not present during the performance;
and a person shall not be treated as aiding or abetting the commission of an offence under this section by reason only of his taking part in a performance as a performer.
(4) In this section “play” and “public performance” have the same meaning as in the Theatres Act 1968.
(5) The following provisions of the Theatres Act 1968 apply in relation to an offence under this section as they apply to an offence under section 2 of that Act—
section 9 (script as evidence of what was performed),
section 10 (power to make copies of script),
section 15 (powers of entry and inspection).
Section 29E Distributing, showing or playing a recording
(1) A person who distributes, or shows or plays, a recording of visual images or sounds which are threatening is guilty of an offence if he intends thereby to stir up religious hatred.
(2) In this Part “recording” means any record from which visual images or sounds may, by any means, be reproduced; and references to the distribution, showing or playing of a recording are to its distribution, showing or playing to the public or a section of the public.
(3) This section does not apply to the showing or playing of a recording solely for the purpose of enabling the recording to be included in a programme service.
Section 29F Broadcasting or including programme in programme service
(1) If a programme involving threatening visual images or sounds is included in a programme service, each of the persons mentioned in subsection (2) is guilty of an offence if he intends thereby to stir up religious hatred.
(2) The persons are—
(a)the person providing the programme service,
(b)any person by whom the programme is produced or directed, and
(c)any person by whom offending words or behaviour are used.
Inflammatory material
Section 29G Possession of inflammatory material
(1) A person who has in his possession written material which is threatening, or a recording of visual images or sounds which are threatening, with a view to—
(a)in the case of written material, its being displayed, published, distributed, or included in a programme service whether by himself or another, or
(b)in the case of a recording, its being distributed, shown, played, or included in a programme service, whether by himself or another,
is guilty of an offence if he intends religious hatred to be stirred up thereby.
(2) For this purpose regard shall be had to such display, publication, distribution, showing, playing, or inclusion in a programme service as he has, or it may reasonably be inferred that he has, in view.
Section 29H: Powers of entry and search
(1) If in England and Wales a justice of the peace is satisfied by information on oath laid by a constable that there are reasonable grounds for suspecting that a person has possession of written material or a recording in contravention of section 29G, the justice may issue a warrant under his hand authorising any constable to enter and search the premises where it is suspected the material or recording is situated.
(2) If in Scotland a sheriff or justice of the peace is satisfied by evidence on oath that there are reasonable grounds for suspecting that a person has possession of written material or a recording in contravention of section 29G, the sheriff or justice may issue a warrant authorising any constable to enter and search the premises where it is suspected the material or recording is situated.
(3) A constable entering or searching premises in pursuance of a warrant issued under this section may use reasonable force if necessary.
(4) In this section “premises” means any place and, in particular, includes—
(a)any vehicle, vessel, aircraft or hovercraft,
(b)any offshore installation as defined in section 12 of the Mineral Workings (Offshore Installations) Act 1971, and
(c)any tent or movable structure.
Section 29I: Power to order forfeiture
(1) A court by or before which a person is convicted of—
(a)an offence under section 29B relating to the display of written material, or
(b)an offence under section 29C, 29E or 29G,
shall order to be forfeited any written material or recording produced to the court and shown to its satisfaction to be written material or a recording to which the offence relates.
(2) An order made under this section shall not take effect—
(a)in the case of an order made in proceedings in England and Wales, until the expiry of the ordinary time within which an appeal may be instituted or, where an appeal is duly instituted, until it is finally decided or abandoned;
(b)in the case of an order made in proceedings in Scotland, until the expiration of the time within which, by virtue of any statute, an appeal may be instituted or, where such an appeal is duly instituted, until the appeal is finally decided or abandoned.
(3) For the purposes of subsection (2)(a)—
(a)an application for a case stated or for leave to appeal shall be treated as the institution of an appeal, and
(b)where a decision on appeal is subject to a further appeal, the appeal is not finally determined until the expiry of the ordinary time within which a further appeal may be instituted or, where a further appeal is duly instituted, until the further appeal is finally decided or abandoned.
(4) For the purposes of subsection (2)(b) the lodging of an application for a stated case or note of appeal against sentence shall be treated as the institution of an appeal.
Section 29J Protection of freedom of expression
Nothing in this Part shall be read or given effect in a way which prohibits or restricts discussion, criticism or expressions of antipathy, dislike, ridicule, insult or abuse of particular religions or the beliefs or practices of their adherents, or of any other belief system or the beliefs or practices of its adherents, or proselytising or urging adherents of a different religion or belief system to cease practising their religion or belief system.

\section{Annotation Guidelines}
\label{app:guidelines}

The guidelines provided for the annotation are reproduced next.

\begin{footnotesize}
\begin{tabular}{|p{0.9\columnwidth}|}
\hline 
\bf Hate speech laws annotation \\
You are asked to annotate 100
instances according to 3 different hate speech laws.
Please find the sentences in the separate excel file. All of these sentences are originally
annotated as hate speech; however, we want to know how many of these can be
prosecutable hate speech under three different laws: Greek, Italian, UK. The laws are
taken from this website. You can find the laws down below. Your job is to annotate in the
designated column whether an instance is \textbf{prosecutable} hate speech by assigning the
value 1 or not by assigning the value 0. Feel free to use the comments column for any
thoughts. If you have any questions contact\ldots   \\
\hline 
\end{tabular}
\begin{tabular}{llcccl}
\hline 
        &           & \multicolumn{3}{c}{\bf Legal context} \\
\bf id  & \bf Text  & \bf Greek & \bf Italian & \bf UK & \bf comments    \\  
\hline 
Id1  & Instance1 & 0 & 1 & 1    \\
Id2  & Instance2 & 1 & 0 & 1    \\ 
\ldots  \\
\hline 
\end{tabular}
\end{footnotesize}

\section{Prompts}
\label{appendixC}

Here we include the prompts used with chatGPT 3.5 for each of the approaches.

\noindent
\textbf{0-shot\_g}

\noindent
Is the following sentence prosecutable hate speech according? Reply with JUST a number:(0) not prosecutable (1) unlikely prosecutable (2) likely prosecutable (3) surely prosecutable. {[}sentence{]}



\noindent
\textbf{0-shot w/Law }

\noindent
Is the following sentence prosecutable hate speech according to the {[}country name{]} law? Reply with JUST a number:(0) not prosecutable (1) unlikely prosecutable (2) likely prosecutable (3) surely prosecutable. {[}country law{]} {[}sentence{]}

\noindent
\textbf{few-shot \& LOOCV}

\noindent
Is the following sentence prosecutable hate speech according to the examples? Reply with JUST a number:(0) not prosecutable (1) unlikely prosecutable (2) likely prosecutable (3) surely prosecutable.{[}examples{{]}{[}sentence{]} 

\noindent
\textbf{few-shot \& LOOCV w/Law}

\noindent
Is the following sentence prosecutable hate speech according to the examples and the law? Reply with JUST a number:(0) not prosecutable (1) unlikely prosecutable (2) likely prosecutable (3) surely prosecutable.{[}law{]} {[}examples{{]}{[}sentence{]} 

\section{Multilabel Error Scores and Multiclass F$_1$ Scores}
\label{appendixD}
\FloatBarrier 
\begin{table*}[htbp]
    \centering
    \scriptsize
    \begin{tabular}{@{}lcccccccccccc@{}}
        & \multicolumn{3}{c}{\textbf{HateBERT}} & \multicolumn{3}{c}{\textbf{DehateBERT}} & \multicolumn{3}{c}{\textbf{HateRoBERTa}} & \multicolumn{3}{c}{\textbf{LegalBERT}} \\
        \cmidrule(lr){2-4} \cmidrule(lr){5-7} \cmidrule(lr){8-10} \cmidrule(lr){11-13}
        \textbf{Country}   & Micro & Macro & Weighted & Micro & Macro & Weighted & Micro & Macro & Weighted & Micro & Macro & Weighted\\
        \midrule
        Greece  & 0.69 & 0.47 & 0.66 & 0.55 & 0.38 & 0.47 & 0.58 & 0.37 & 0.54 & 0.50 & 0.29 & 0.46 \\
        Italy & 0.75 & 0.47 & 0.70 & 0.64 & 0.52 & 0.34 & 0.69 & 0.53 & 0.66 & 0.64 & 0.37  & 0.60 \\
        UK & 0.72 & 0.53 & 0.67 & 0.67 & 0.37 & 0.60 & 0.69 & 0.45 & 0.66 & 0.67 & 0.37 & 0.63  \\
        \bottomrule
    \end{tabular}%
    \caption{F$_1$ scores of PLMs.}
    \label{tab:plms_f1_1}
\end{table*}

\begin{table*}
    \centering
    \scriptsize
    \begin{tabular}{@{}lcccccccccccc@{}}
        & \multicolumn{3}{c}{\textbf{HateBERT}} & \multicolumn{3}{c}{\textbf{DehateBERT}} & \multicolumn{3}{c}{\textbf{HateRoBERTa}} & \multicolumn{3}{c}{\textbf{LegalBERT}} \\
        \cmidrule(lr){2-4} \cmidrule(lr){5-7} \cmidrule(lr){8-10} \cmidrule(lr){11-13}
        \textbf{Country}   & Micro & Macro & Weighted & Micro & Macro & Weighted & Micro & Macro & Weighted & Micro & Macro & Weighted\\
        \midrule
        Greece  & 0.32 & 0.32 & 0.19 &  0.14  & 0.11 & 0.12 & 0.42 & 0.24 & 0.36 & 0.12 & 0.08 & 0.12 \\
        Italy & 0.20 & 0.12 & 0.20 & 0.13 & 0.10 & 0.11 & 0.44 & 0.21 & 0.40 & 0.13 & 0.07  & 0.08 \\
        UK & 0.24 & 0.15 & 0.27 & 0.17 & 0.11 & 0.18 & 0.59 & 0.19 & 0.44 & 0.11 & 0.06 & 0.08  \\
        \bottomrule
    \end{tabular}%
    \caption{F$_1$ scores of PLMs per class when trained on the silver labels.}
    \label{tab:plms_f1_2}
\end{table*}


\begin{table*}
\small
    \centering
    \resizebox{\textwidth}{!}{
    \begin{tabular}{@{}lcccccc|cccccc@{}}
       & \multicolumn{3}{c}{\textbf{0-shot}} & \multicolumn{3}{c}{\textbf{0-shot w/Law}} & \multicolumn{3}{c}{\textbf{LOOCV few-shot}} & \multicolumn{3}{c}{\textbf{LOOCV few-shot w/Law}} \\ \cmidrule(lr){2-4} \cmidrule(lr){5-7} \cmidrule(lr){8-10} \cmidrule(lr){11-13}
     \textbf{Country}   & Micro & Macro & Weighted & Micro & Macro & Weighted & Micro & Macro & Weighted & Micro & Macro & Weighted\\
     \hline
       Greece  & 0.26 & 0.26 & 0.24 & 0.37 & 0.28 & 0.34 & 0.52 & 0.31 & 0.58 & 0.53 & 0.35 & 0.57\\
       Italy & 0.25 & 0.24 & 0.24 & 0.37 & 0.29 & 0.31 & 0.59 & 0.28 & 0.70 & 0.56 & 0.27 & 0.64\\
       UK & 0.23 & 0.20 & 0.21 & 0.21 & 0.18 & 0.19 & 0.61 & 0.30 & 0.69 & 0.55 & 0.37 & 0.56\\
       \bottomrule
    \end{tabular}%
    }

    \caption{F$_1$ scores per class for Qwen2 for 0-shot and LOOCV settings.}    
    \label{tab:zero_loo_gpt_f1} 
\end{table*}

\begin{table*}
\small
    \centering
    \resizebox{\textwidth}{!}{
    \begin{tabular}{@{}lcccccc|cccccc@{}}
       & \multicolumn{3}{c}{\textbf{0-shot}} & \multicolumn{3}{c}{\textbf{0-shot w/Law}} & \multicolumn{3}{c}{\textbf{LOOCV few-shot}} & \multicolumn{3}{c}{\textbf{LOOCV few-shot w/Law}} \\ \cmidrule(lr){2-4} \cmidrule(lr){5-7} \cmidrule(lr){8-10} \cmidrule(lr){11-13}
     \textbf{Country}   & Micro & Macro & Weighted & Micro & Macro & Weighted & Micro & Macro & Weighted & Micro & Macro & Weighted\\
     \hline
       Greece  & 0.24 & 0.22 & 0.24 & 0.16 & 0.15 & 0.13 & 0.31 & 0.24 & 0.31 & 0.24 & 0.16 & 0.26\\
       Italy & 0.13 & 0.13 & 0.13 & 0.16 & 0.13 & 0.17 & 0.37 & 0.32 & 0.32 & 0.33 & 0.29 & 0.30\\
       UK & 0.20 & 0.21 & 0.18 & 0.28 & 0.22 & 0.26 & 0.35 & 0.33 & 0.32 & 0.32 & 0.30 & 0.30\\
       \bottomrule
    \end{tabular}%
    }

    \caption{F$_1$ scores per class for Llama3 for 0-shot and LOOCV settings.}    
    \label{tab:mistral_f1_zero_loo} 
\end{table*}


\begin{table*}
    \centering
    \small
    \begin{tabular}{@{}lccccccccc@{}}
        & \multicolumn{3}{c}{\textbf{4-shot}} & \multicolumn{3}{c}{\textbf{8-shot}} & \multicolumn{3}{c}{\textbf{12-shot}} \\
        \cmidrule(lr){2-4} \cmidrule(lr){5-7} \cmidrule(lr){8-10} 
        \textbf{Country} & Micro & Macro & Weighted & Micro & Macro & Weighted & Micro & Macro & Weighted  \\
        \midrule
        Greece  & 0.37 & 0.29 & 0.36 & 0.41 & 0.25 & 0.45 & 0.39 & 0.32 & 0.43\\
        Italy & 0.41 & 0.30 & 0.39 & 0.46 & 0.33 & 0.45 & 0.59 & 0.40 & 0.62  \\
        UK & 0.47 & 0.36 & 0.43 & 0.48 & 0.22 & 0.51 & 0.52 & 0.34 & 0.54 \\
        \bottomrule
    \end{tabular}%
    \caption{F$_1$ scores of Qwen2 in the few-shot setting without law.}
    \label{tab:few_shots_f1_gpt_no_law}
\end{table*}

\begin{table*}
    \centering
    \small
    \begin{tabular}{@{}lccccccccc@{}}
        & \multicolumn{3}{c}{\textbf{4-shot}} & \multicolumn{3}{c}{\textbf{8-shot}} & \multicolumn{3}{c}{\textbf{12-shot}} \\
        \cmidrule(lr){2-4} \cmidrule(lr){5-7} \cmidrule(lr){8-10} 
        \textbf{Country}   & Micro & Macro & Weighted & Micro & Macro & Weighted & Micro & Macro & Weighted  \\
        \midrule
        Greece  & 0.35 & 0.27 & 0.31 & 0.30 & 0.26 & 0.26 & 0.34 & 0.29 & 0.32\\
        Italy & 0.33 & 0.27 & 0.30 & 0.28 & 0.22 & 0.24 & 0.28  & 0.21 & 0.26  \\
        UK & 0.41 & 0.35 & 0.39 & 0.32 & 0.31 & 0.34 & 0.34 & 0.30 & 0.29 \\
        \bottomrule
    \end{tabular}%
    \caption{F$_1$ scores of Llama3 in the few-shot setting without law.}
    \label{tab:few_shots_f1_mistral}
\end{table*}


\begin{table*}
    \centering
    \small
    \begin{tabular}{@{}lccccccccc@{}}
        & \multicolumn{3}{c}{\textbf{4-shot}} & \multicolumn{3}{c}{\textbf{8-shot}} & \multicolumn{3}{c}{\textbf{12-shot}} \\
        \cmidrule(lr){2-4} \cmidrule(lr){5-7} \cmidrule(lr){8-10} 
        \textbf{Country}   & Micro & Macro & Weighted & Micro & Macro & Weighted & Micro & Macro & Weighted  \\
        \midrule
        Greece  & 0.42 & 0.33 & 0.41 & 0.41 & 0.34 & 0.44 & 0.38 & 0.29 & 0.39\\
        Italy & 0.59 & 0.39 & 0.56 & 0.53 & 0.37 & 0.53 & 0.58  & 0.45 & 0.58  \\
        UK & 0.45 & 0.35 & 0.40 & 0.42 & 0.28 & 0.37 & 0.51 & 0.37 & 0.48 \\
        \bottomrule
    \end{tabular}%
    \caption{F$_1$ scores of Qwen2 in the few-shot setting with law.}
    \label{tab:few_shots_f1_gpt_w/law}
\end{table*}


\begin{table*}
    \centering
    \small
    \begin{tabular}{@{}lccccccccc@{}}
        & \multicolumn{3}{c}{\textbf{4-shot}} & \multicolumn{3}{c}{\textbf{8-shot}} & \multicolumn{3}{c}{\textbf{12-shot}} \\
        \cmidrule(lr){2-4} \cmidrule(lr){5-7} \cmidrule(lr){8-10} 
        \textbf{Country}   & Micro & Macro & Weighted & Micro & Macro & Weighted & Micro & Macro & Weighted  \\
        \midrule
        Greece  & 0.29 & 0.26 & 0.26 & 0.34 & 0.31 & 0.31 & 0.26 & 0.25 & 0.26\\
        Italy & 0.33 & 0.25 & 0.30 & 0.30 & 0.28 & 0.28 & 0.34 & 0.30 & 0.30  \\
        UK & 0.37 & 0.32 & 0.36 & 0.32 & 0.27 & 0.29 & 0.29 & 0.25 & 0.30 \\
        \bottomrule
    \end{tabular}%
    \caption{F$_1$ scores of Llama3 in the few-shot setting with law.}
    \label{tab:few_shots_f1_mistral_law}
\end{table*}

\section{Confusion Matrices}\label{appendixE}

\FloatBarrier 
\begin{figure*}[htbp]
    \centering
    \includegraphics[width=1\linewidth]{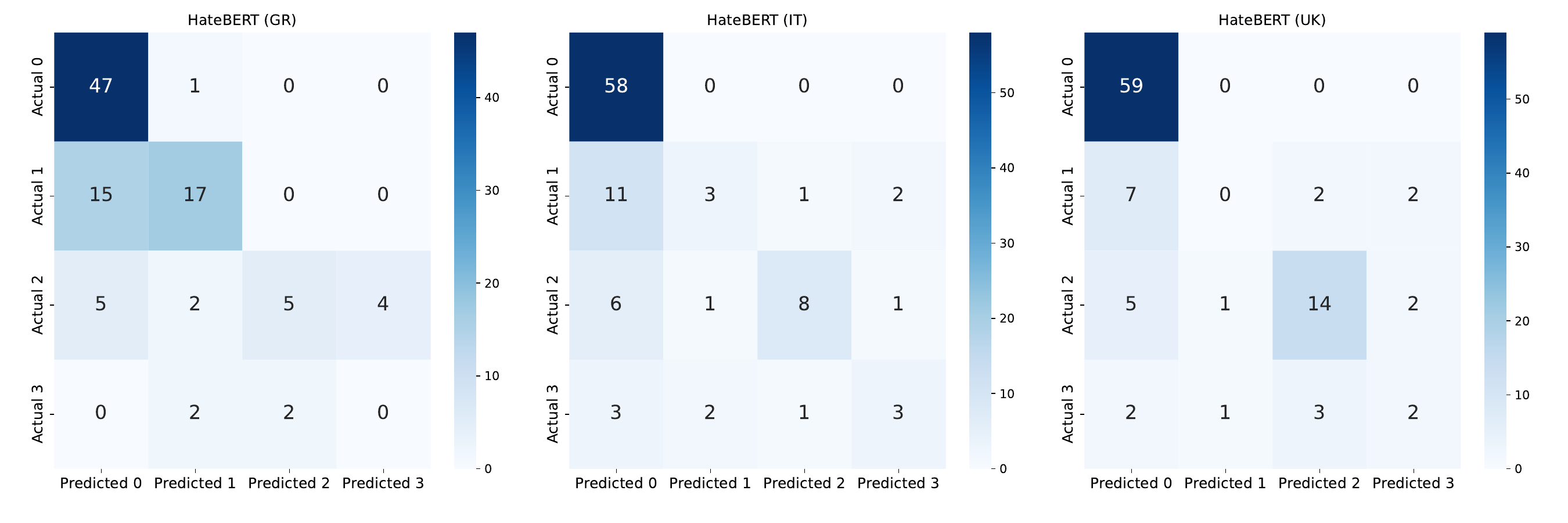}
    \caption{HateBERT confusion matrix.}
    \label{fig:hatebert_cm}
\end{figure*}

\begin{figure*}[htbp]
    \centering
    \includegraphics[width=1\linewidth]{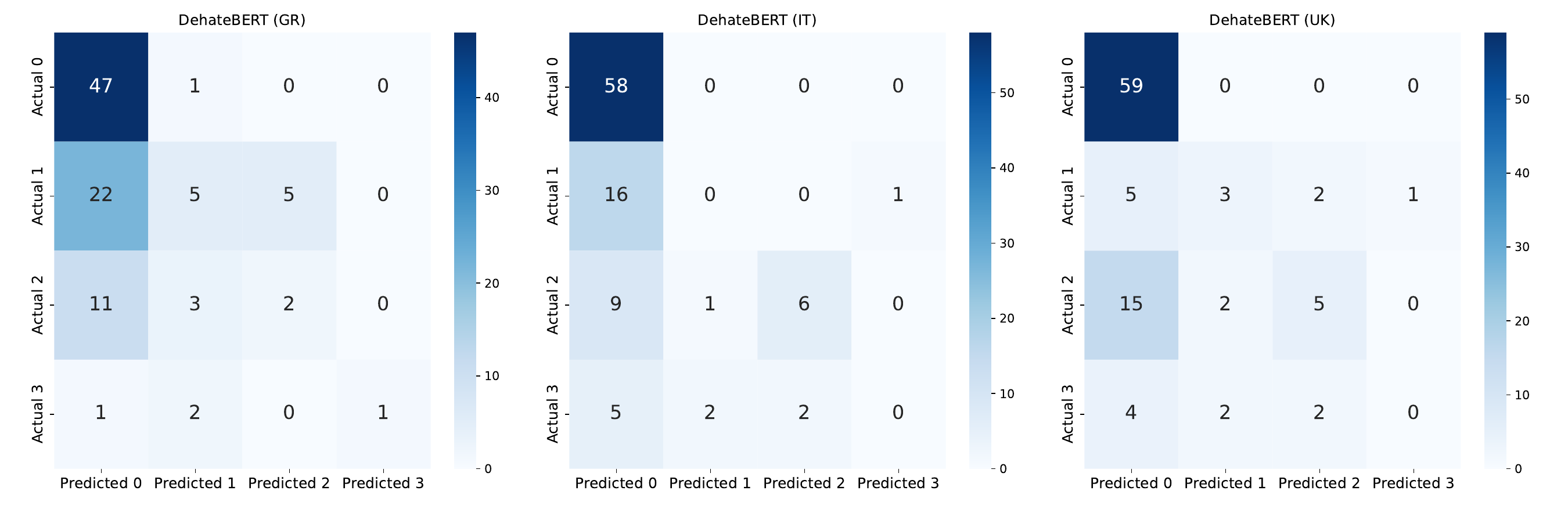}
    \caption{DehateBERT confusion matrix.}
    \label{fig:dehatebert_cm}
\end{figure*}

\begin{figure*}[htbp]
    \centering
    \includegraphics[width=1\linewidth]{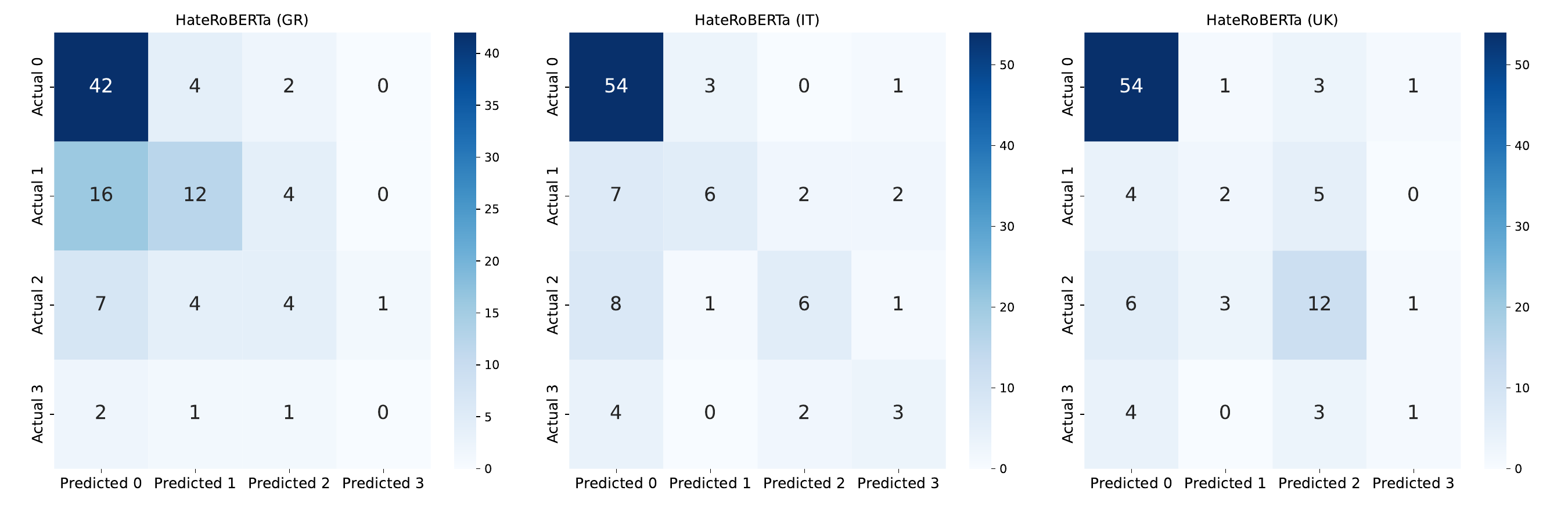}
    \caption{HateRoBERTa confusion matrix.}
    \label{fig:roberta_cm}
\end{figure*}

\begin{figure*}[htbp]
    \centering
    \includegraphics[width=1\linewidth]{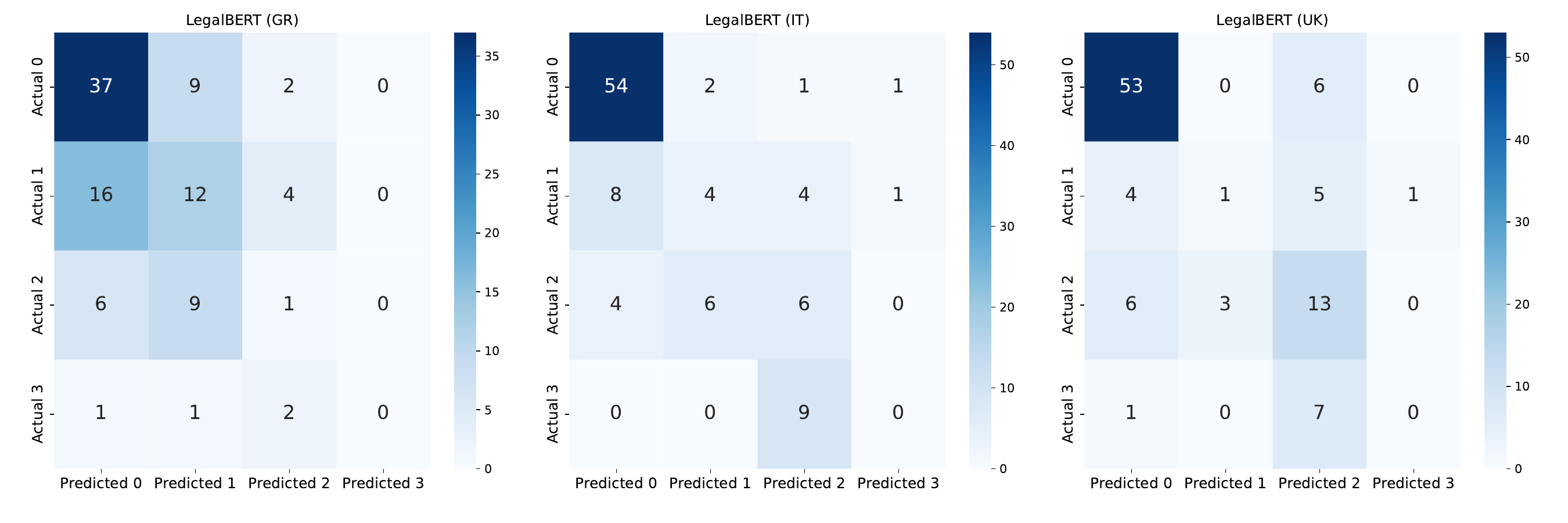}
    \caption{LegalBERT confusion matrix.}
    \label{fig:legalbert_cm}
\end{figure*}

\begin{figure*}[htbp]
    \centering
    \includegraphics[width=1\linewidth]{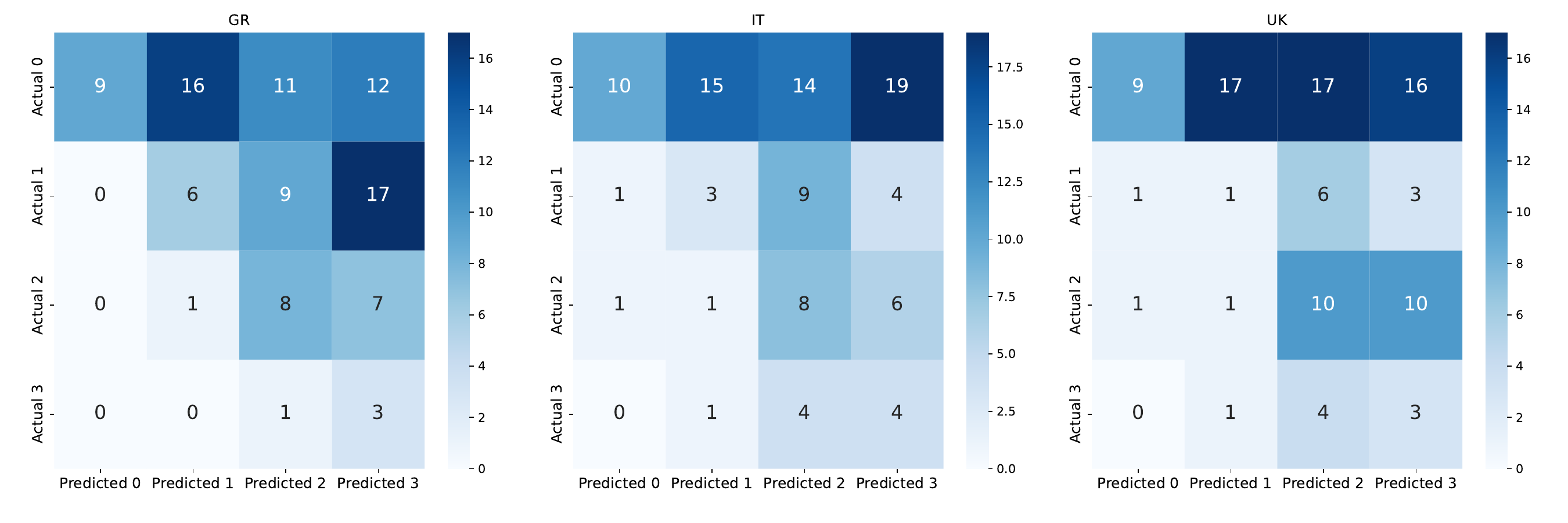}
    \caption{Qwen2 0-shot w/o Law confusion matrix.}
    \label{fig:qwen_0wo_cm}
\end{figure*}

\begin{figure*}[htbp]
    \centering
    \includegraphics[width=1\linewidth]{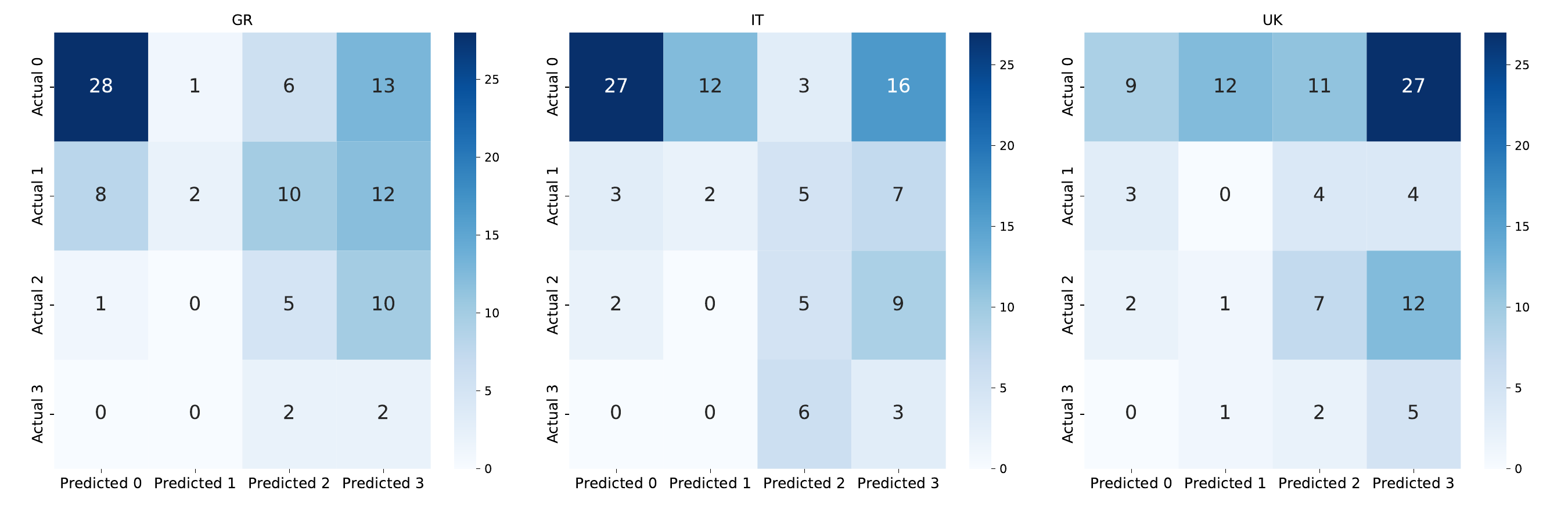}
    \caption{Qwen2 0-shot w/ Law confusion matrix.}
    \label{fig:qwen_0wlaw_cm}
\end{figure*}

\begin{figure*}[htbp]
    \centering
    \includegraphics[width=1\linewidth]{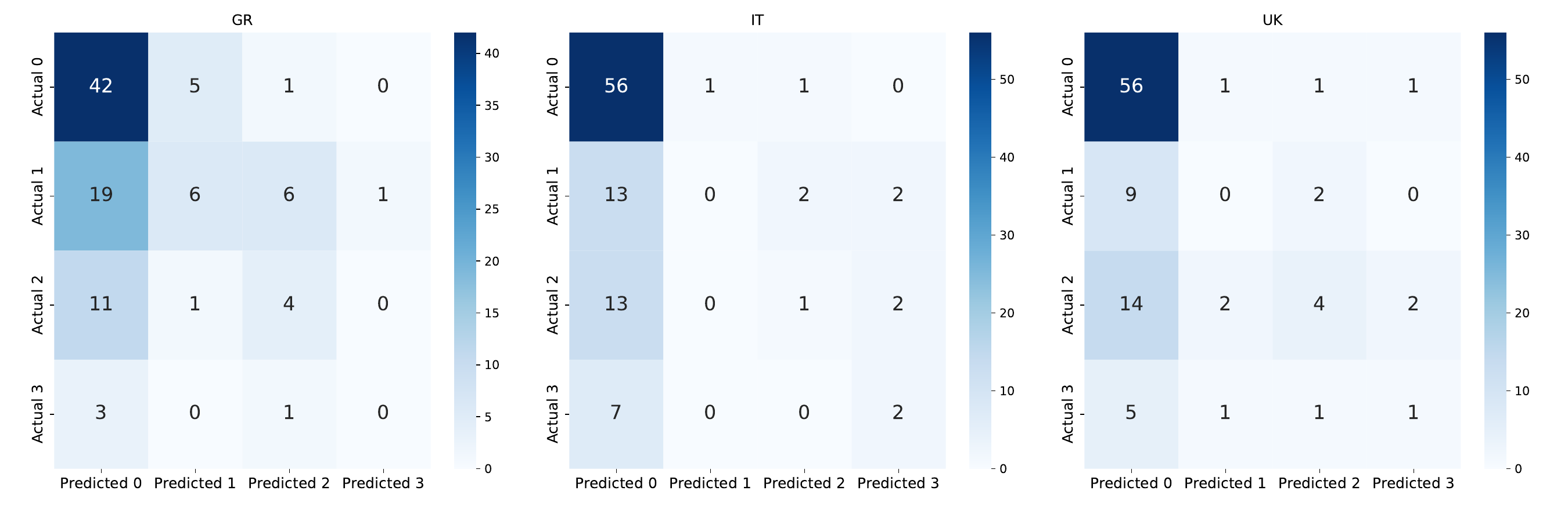}
    \caption{Qwen2 LOOCV w/o Law confusion matrix.}
    \label{fig:qwen_99wo_cm}
\end{figure*}

\begin{figure*}[htbp]
    \centering
    \includegraphics[width=1\linewidth]{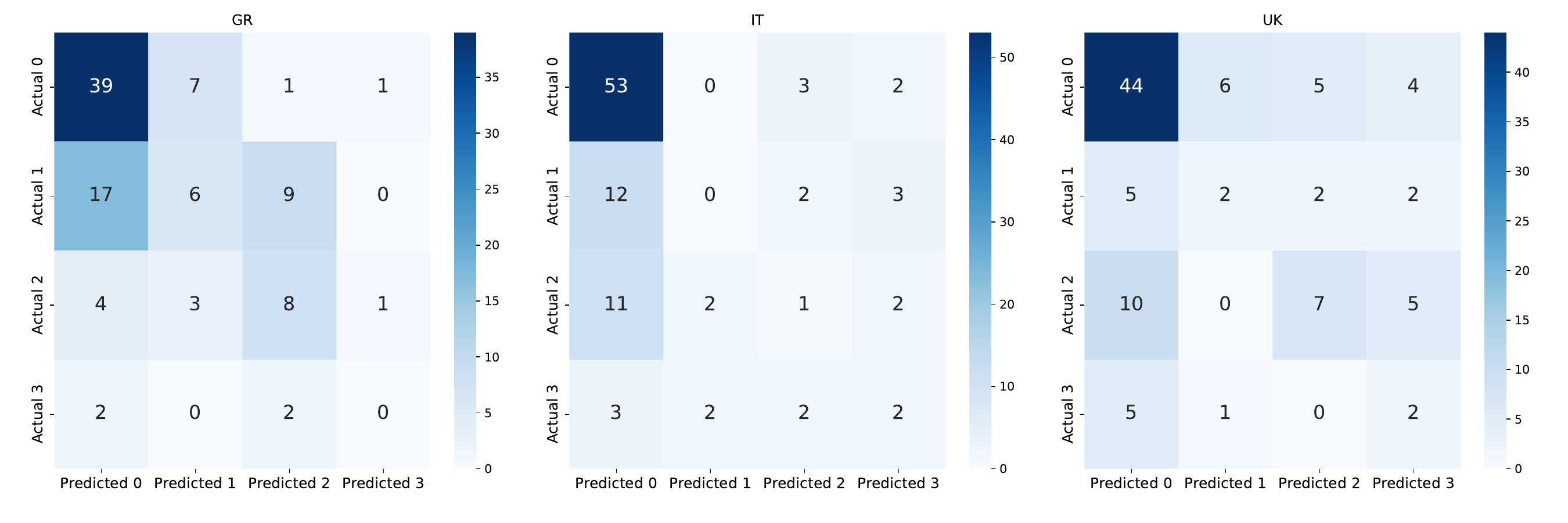}
    \caption{Qwen2 LOOCV w/ Law confusion matrix.}
    \label{fig:qwen_99wlaw_cm}
\end{figure*}

\begin{figure*}[htbp]
    \centering
    \includegraphics[width=1\linewidth]{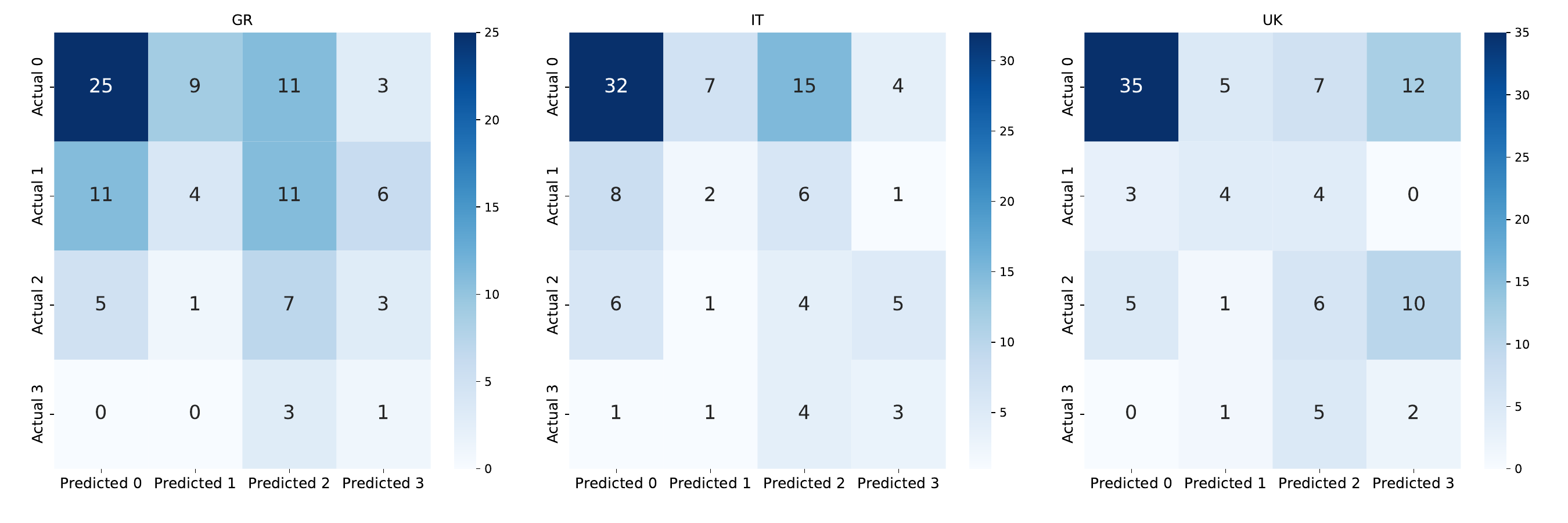}
    \caption{Qwen2 4-shot w/o Law confusion matrix.}
    \label{fig:qwen_4wo_cm}
\end{figure*}

\begin{figure*}[htbp]
    \centering
    \includegraphics[width=1\linewidth]{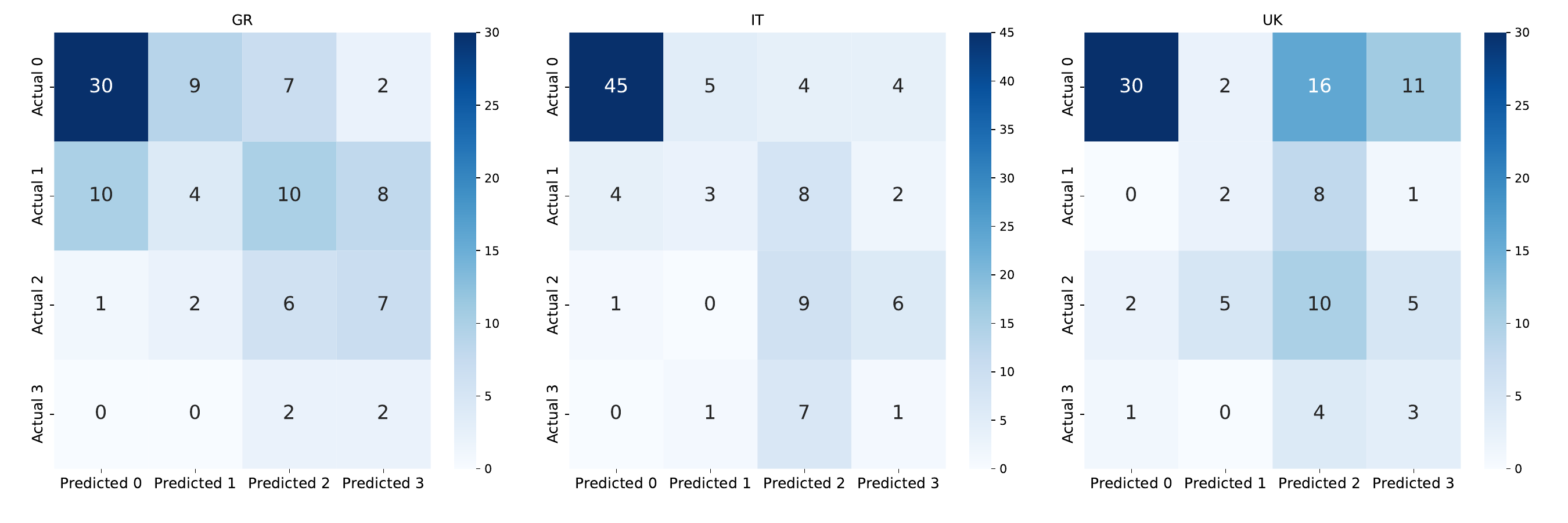}
    \caption{Qwen2 4-shot w/ Law confusion matrix.}
    \label{fig:qwen_4wlaw_cm}
\end{figure*}

\begin{figure*}[htbp]
    \centering
    \includegraphics[width=1\linewidth]{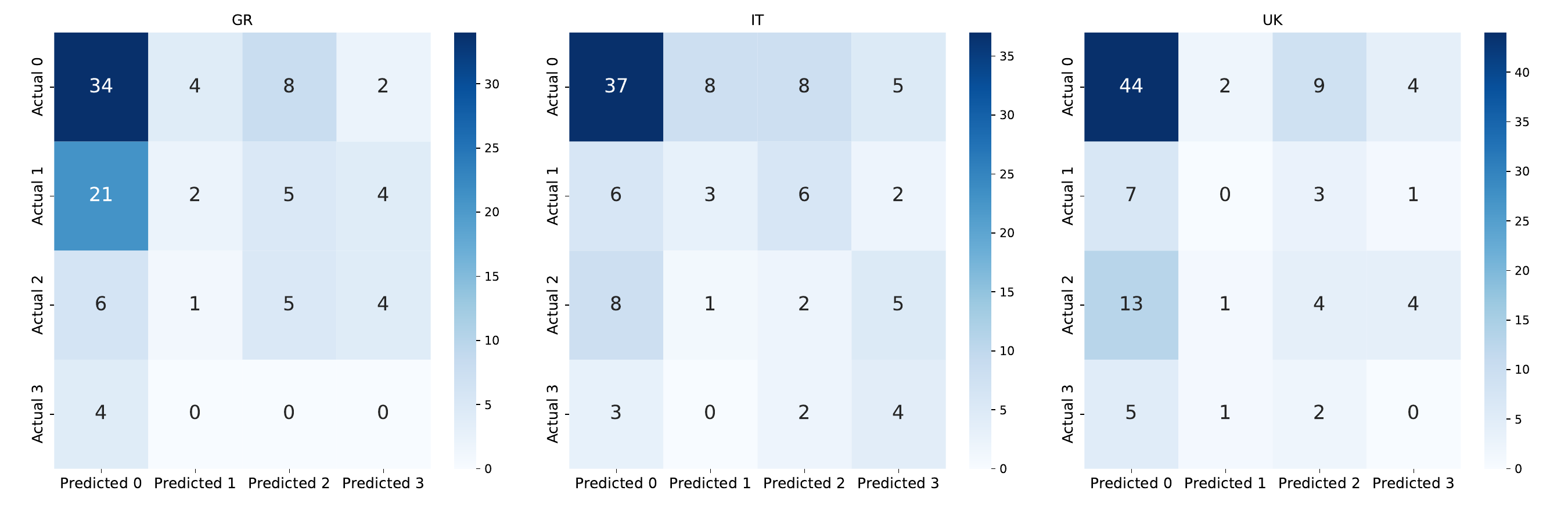}
    \caption{Qwen2 8-shot w/o Law confusion matrix.}
    \label{fig:qwen_8wo_cm}
\end{figure*}

\begin{figure*}[htbp]
    \centering
    \includegraphics[width=1\linewidth]{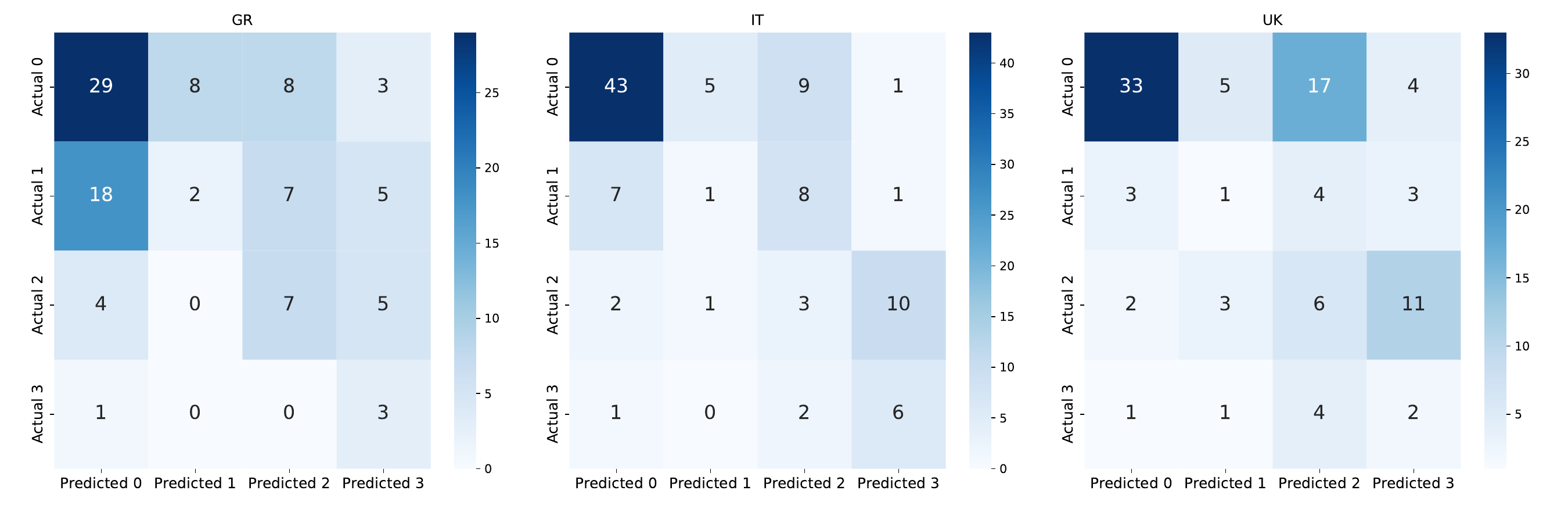}
    \caption{Qwen2 8-shot w/ Law confusion matrix.}
    \label{fig:qwen_8wlaw_cm}
\end{figure*}

\begin{figure*}[htbp]
    \centering
    \includegraphics[width=1\linewidth]{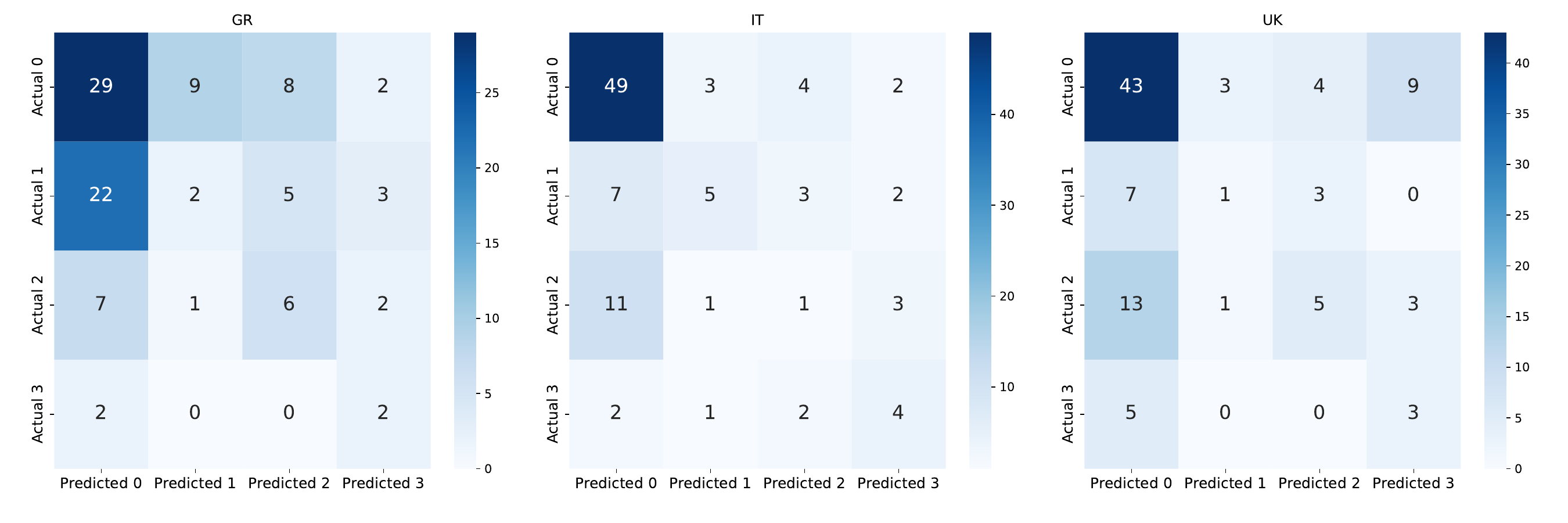}
    \caption{Qwen2 12-shot w/o Law confusion matrix.}
    \label{fig:qwen_12wo_cm}
\end{figure*}

\begin{figure*}[htbp]
    \centering
    \includegraphics[width=1\linewidth]{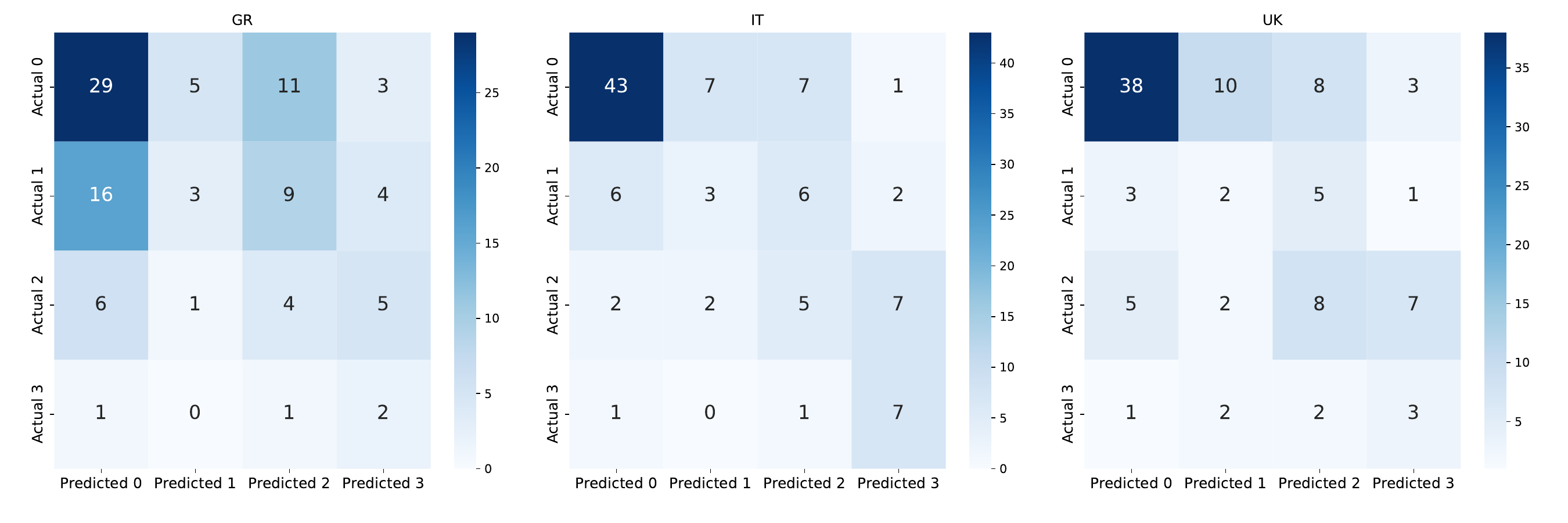}
    \caption{Qwen2 12-shot w/ Law confusion matrix.}
    \label{fig:qwen_12wlaw_cm}
\end{figure*}

\begin{figure*}[htbp]
    \centering
    \includegraphics[width=1\linewidth]{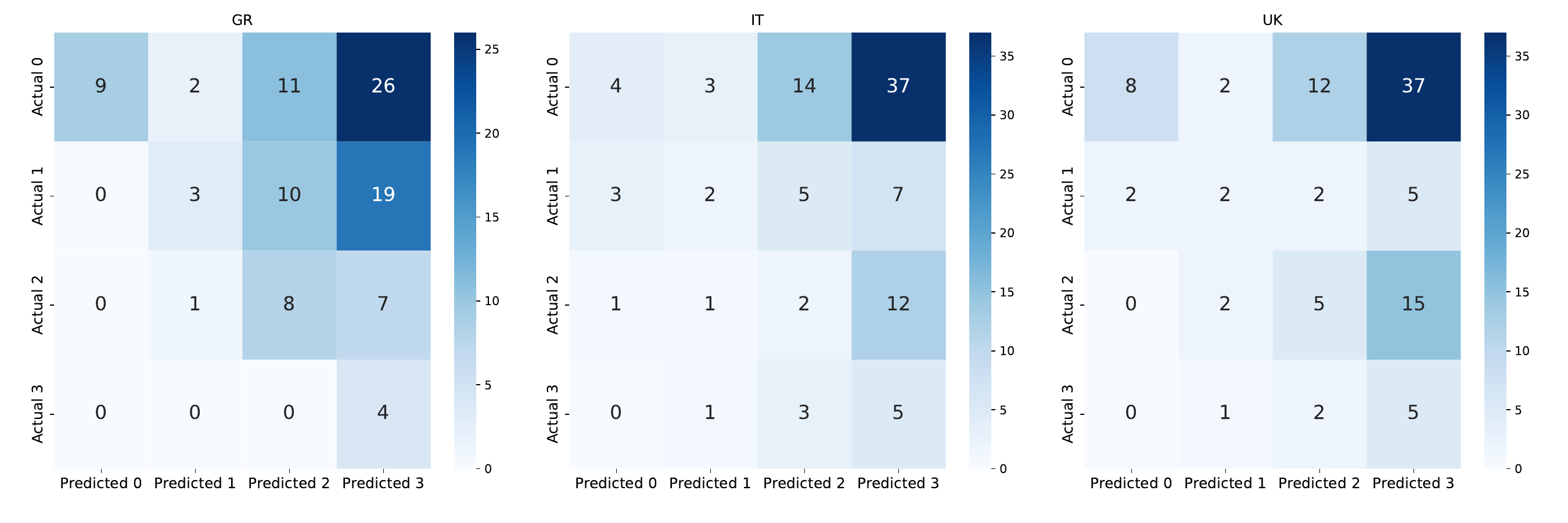}
    \caption{Llama3 0-shot w/o Law confusion matrix.}
    \label{fig:llama_0wo_cm}
\end{figure*}

\begin{figure*}[htbp]
    \centering
    \includegraphics[width=1\linewidth]{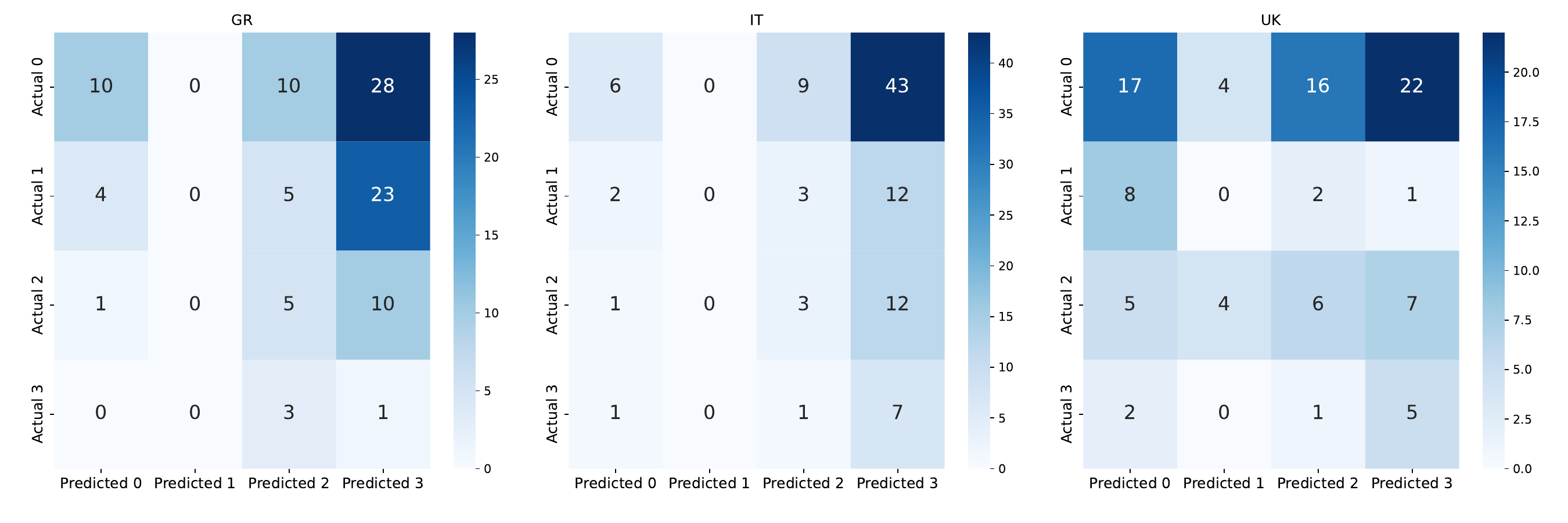}
    \caption{Llama3 0-shot w/ Law confusion matrix.}
    \label{fig:llama_0wlaw_cm}
\end{figure*}

\begin{figure*}[htbp]
    \centering
    \includegraphics[width=1\linewidth]{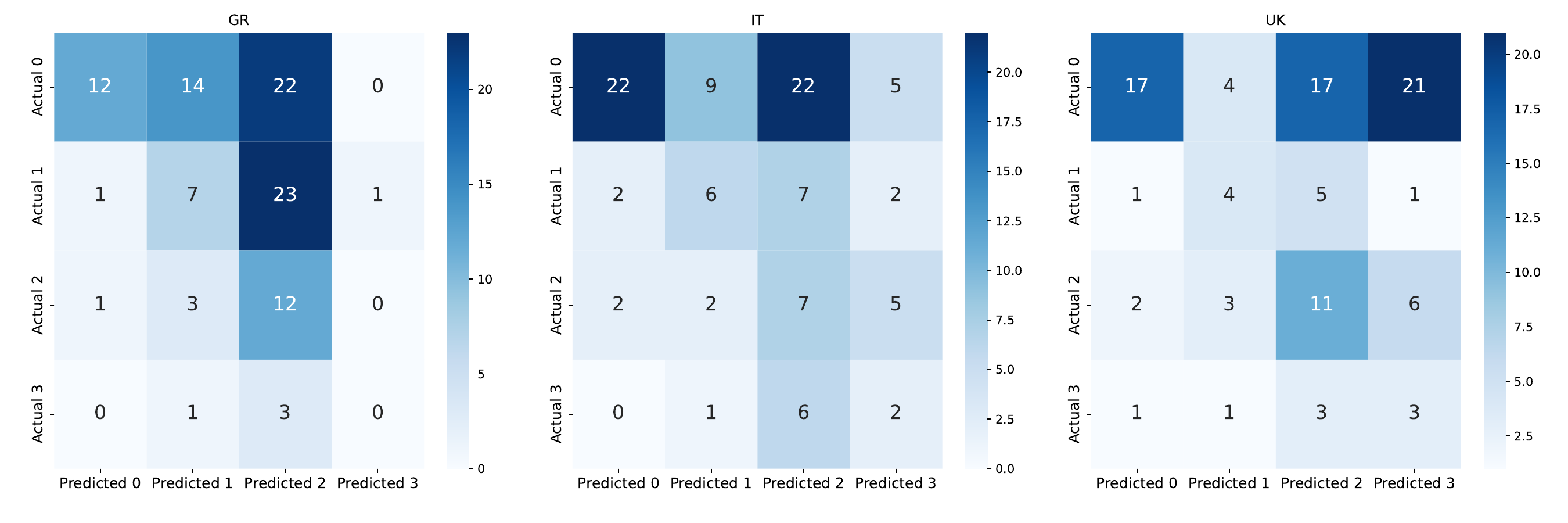}
    \caption{Llama3 LOOCV w/o Law confusion matrix.}
    \label{fig:llama_99wo_cm}
\end{figure*}

\begin{figure*}[htbp]
    \centering
    \includegraphics[width=1\linewidth]{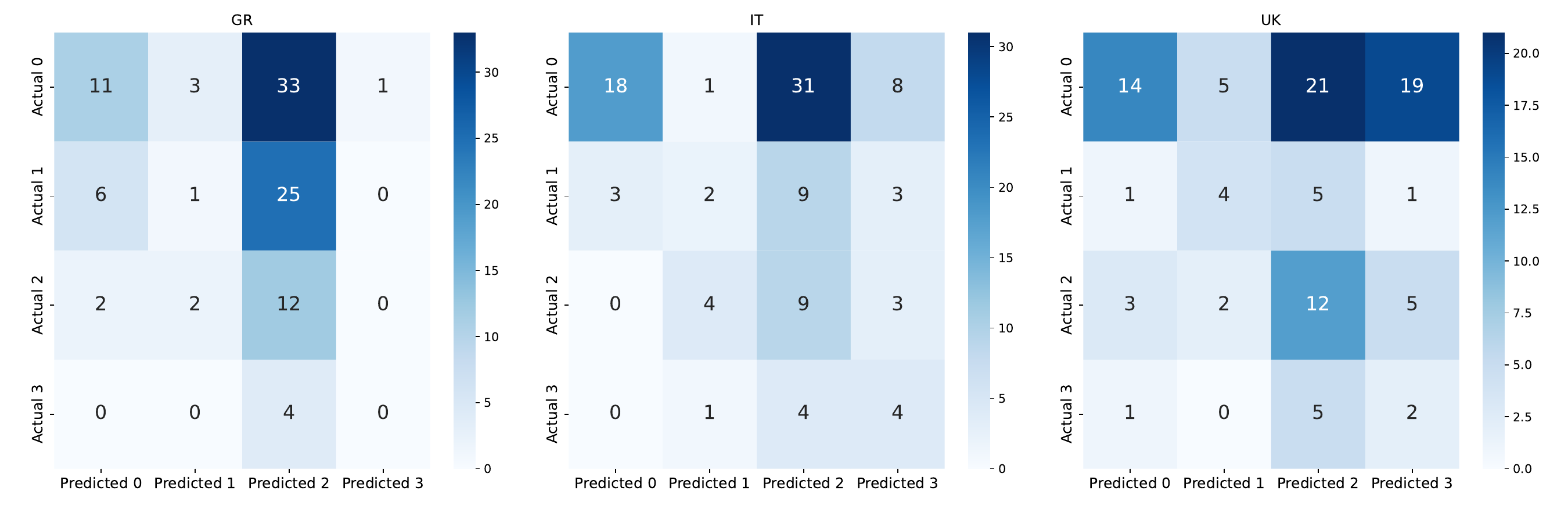}
    \caption{Llama3 LOOCV w/ Law confusion matrix.}
    \label{fig:llama_99wlaw_cm}
\end{figure*}

\begin{figure*}[htbp]
    \centering
    \includegraphics[width=1\linewidth]{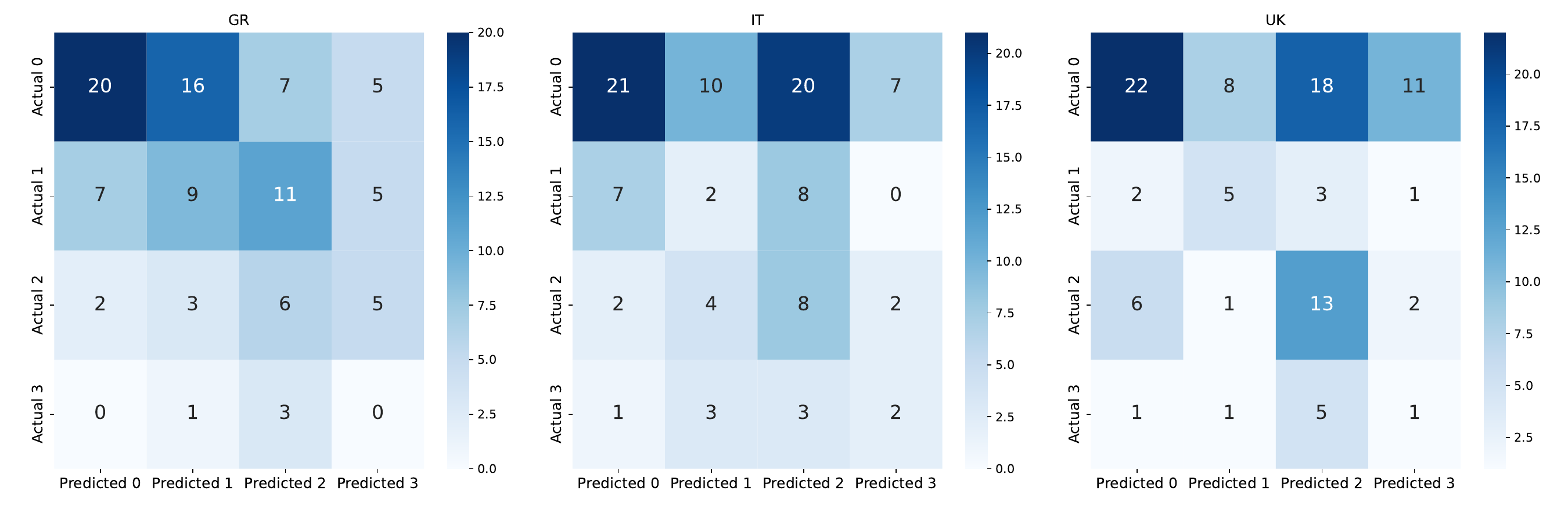}
    \caption{Llama3 4-shot w/o Law confusion matrix.}
    \label{fig:llama_4wo_cm}
\end{figure*}

\begin{figure*}[htbp]
    \centering
    \includegraphics[width=1\linewidth]{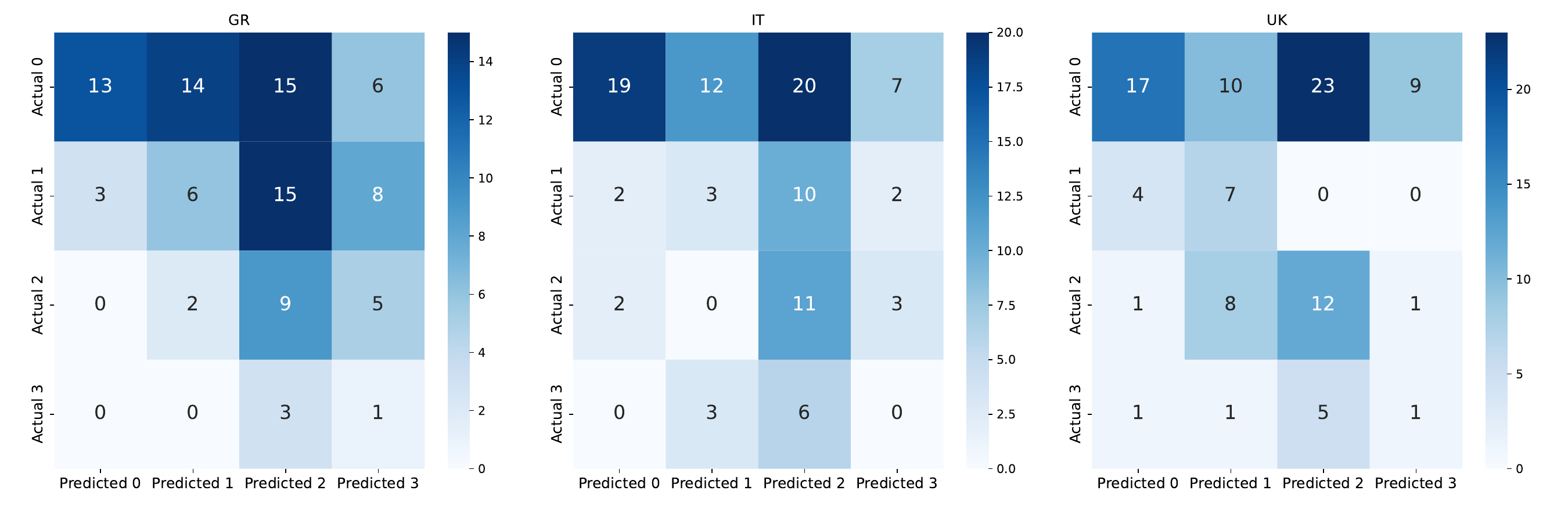}
    \caption{Llama3 4-shot w/ Law confusion matrix.}
    \label{fig:llama_4wlaw_cm}
\end{figure*}

\begin{figure*}[htbp]
    \centering
    \includegraphics[width=1\linewidth]{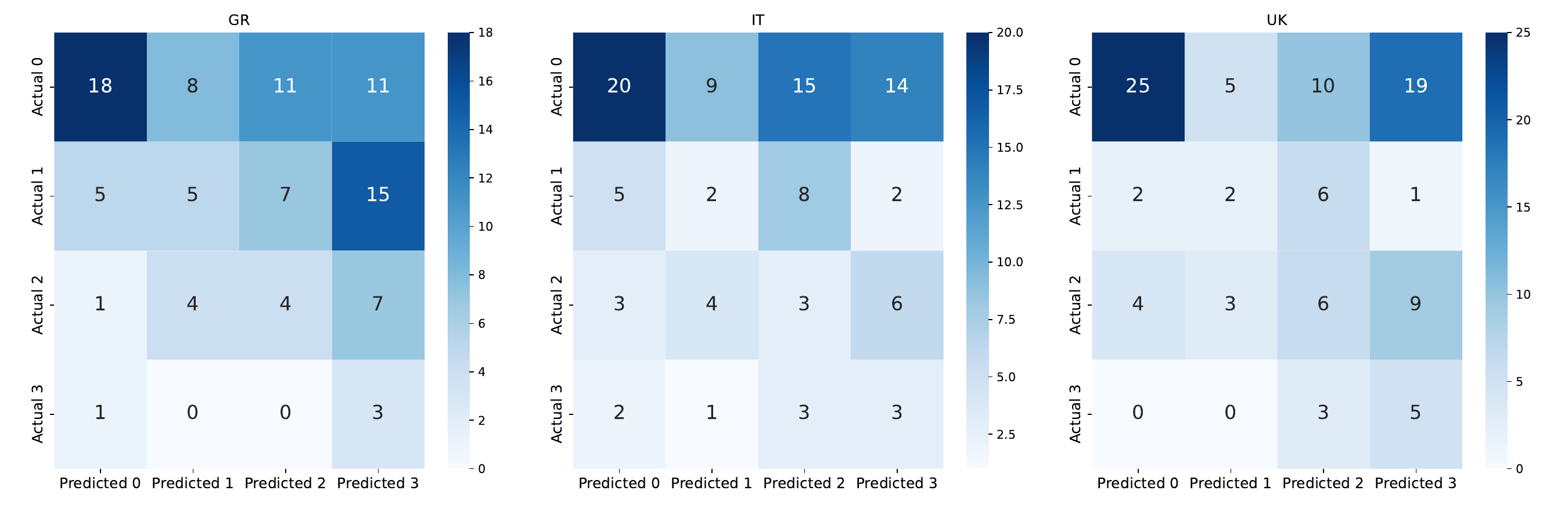}
    \caption{Llama3 8-shot w/o Law confusion matrix.}
    \label{fig:llama_8wo_cm}
\end{figure*}

\begin{figure*}[htbp]
    \centering
    \includegraphics[width=1\linewidth]{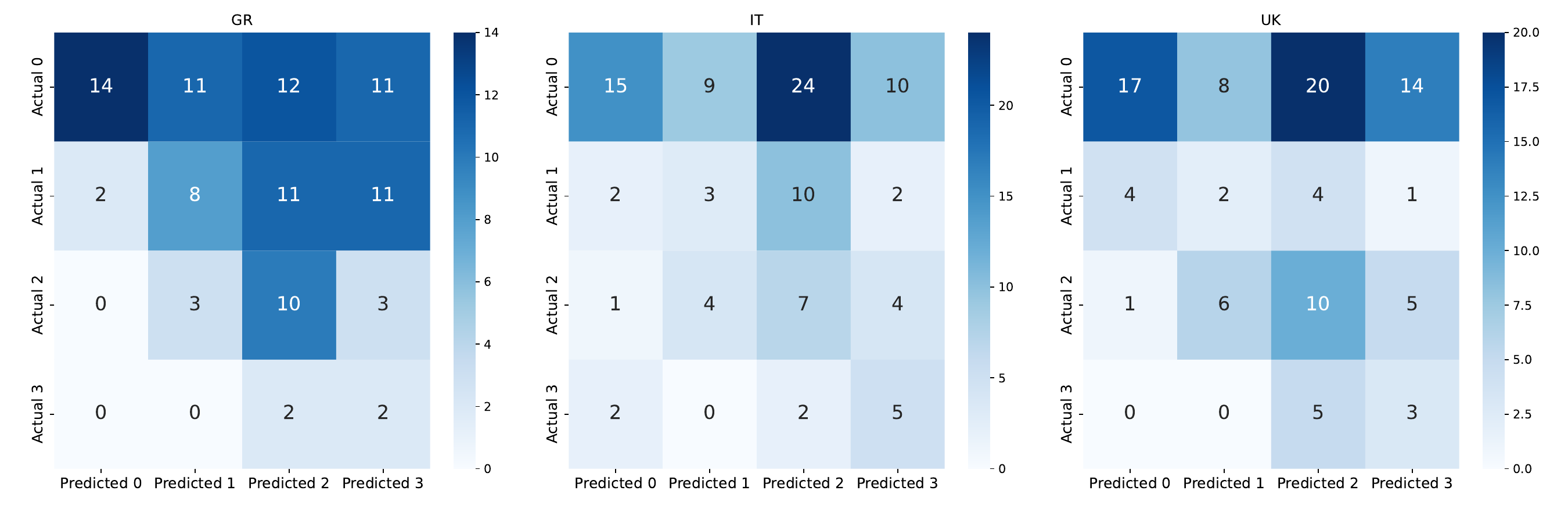}
    \caption{Llama3 8-shot w/ Law confusion matrix.}
    \label{fig:llama_8wlaw_cm}
\end{figure*}

\begin{figure*}[htbp]
    \centering
    \includegraphics[width=1\linewidth]{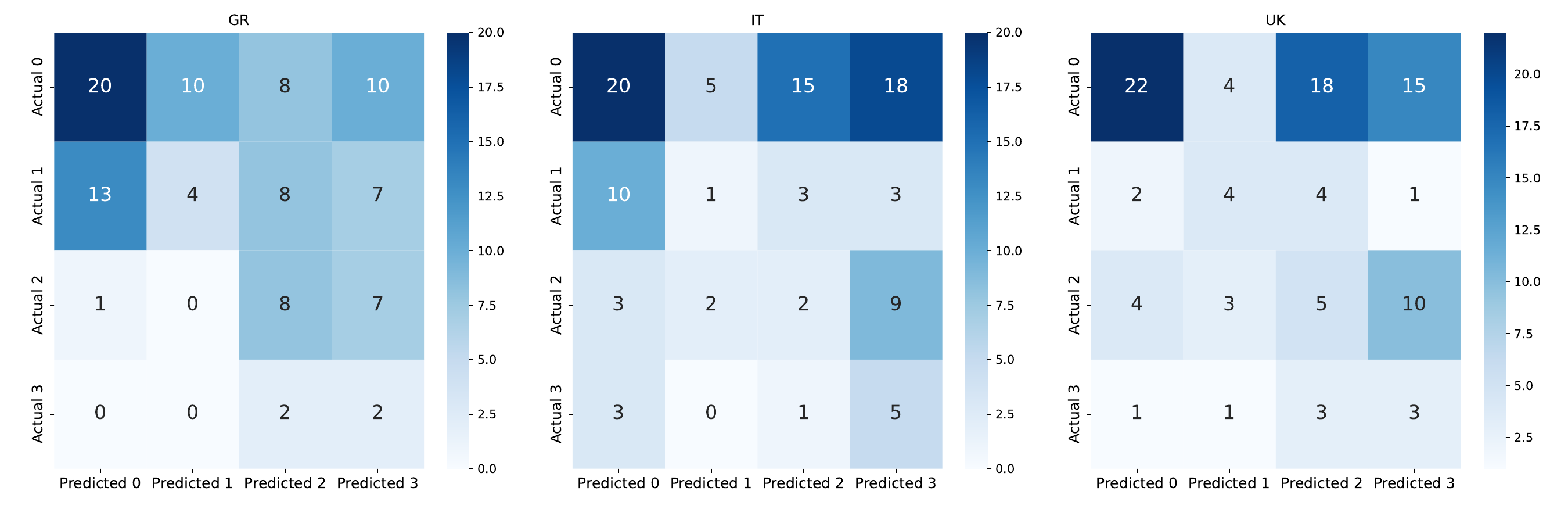}
    \caption{Llama3 12-shot w/o Law confusion matrix.}
    \label{fig:llama_12wo_cm}
\end{figure*}

\begin{figure*}[htbp]
    \centering
    \includegraphics[width=1\linewidth]{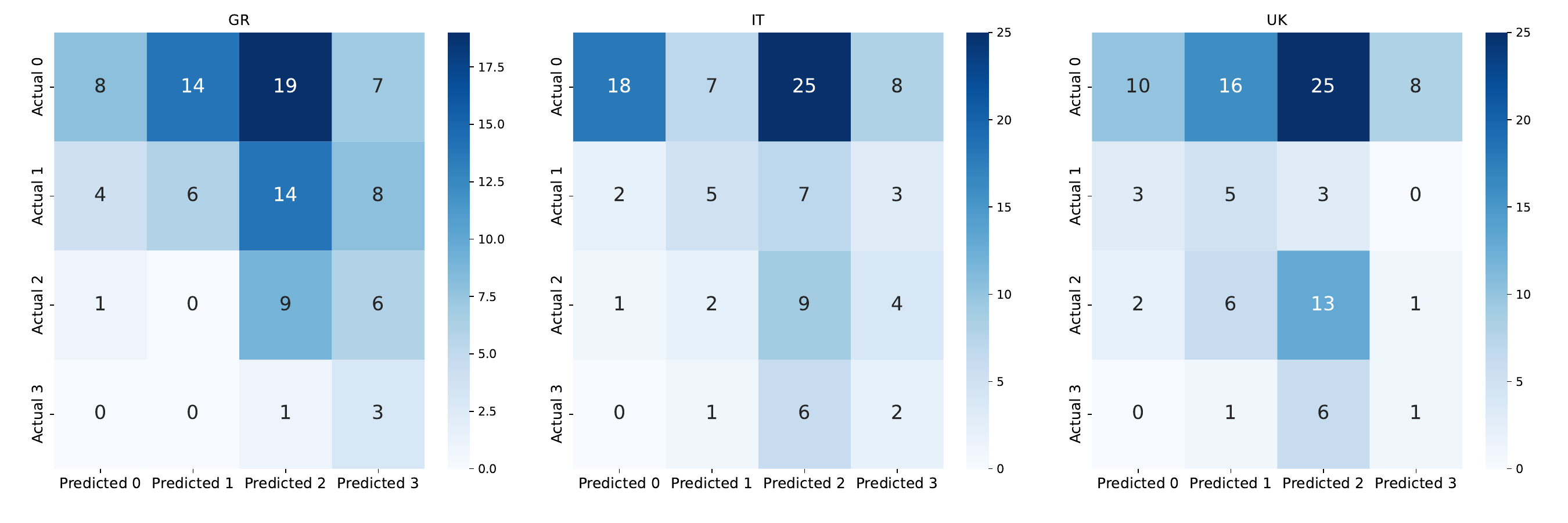}
    \caption{Llama3 12-shot w/ Law confusion matrix.}
    \label{fig:llama_12wlaw_cm}
\end{figure*}
\FloatBarrier

\end{document}